\documentclass[11pt]{article} %
\usepackage[preprint]{mldraft}

\usepackage[utf8]{inputenc} %
\usepackage[T1]{fontenc}    %
\definecolor{MyGreen}{HTML}{009900}
\definecolor{MyBlue}{HTML}{0d3983}
\RequirePackage[colorlinks,citecolor=MyGreen]{hyperref}
\usepackage{url}            %
\usepackage{booktabs}       %
\usepackage{amsfonts}       %
\usepackage{nicefrac}       %
\usepackage{microtype}      %

\usepackage{paperpkg}
\usepackage{subcaption}
\usepackage{textcomp}
\usepackage{soul}

\RequirePackage{algorithm}
\usepackage{etoolbox}

\let\classAND\AND
\let\AND\relax
\usepackage{algorithmic}

\let\AND\classAND
\AtBeginEnvironment{algorithmic}{\let\AND\algoAND}

\usepackage[algo2e,ruled,vlined,linesnumbered]{algorithm2e}
\usepackage{pifont}
\usepackage{booktabs}
\usepackage{upquote}
\usepackage{listings}
\usepackage{comment}
\usepackage{fancyvrb}
\usepackage{graphicx}
\usepackage{multicol}
\usepackage{multirow}
\usepackage{cleveref}
\usepackage{lstautogobble}  %
\usepackage{color}          %
\usepackage{zi4}            %
\usepackage[breakable,skins]{tcolorbox}
\usepackage{listings}
\usepackage{xcolor}
\usepackage{diagbox}
\usepackage{booktabs}
\usepackage{colortbl}
\usepackage{bbding}
\usepackage{graphicx}
\usepackage{rotating}
\usepackage{adjustbox}
\usepackage{lipsum}
\usepackage{enumitem}
\usepackage{framed}

\usepackage{listings}

\newcommand{\matnorm}[1]{{\left\vert\kern-0.25ex\left\vert\kern-0.25ex\left\vert #1 
    \right\vert\kern-0.25ex\right\vert\kern-0.25ex\right\vert}}

\newtheorem{theorem}{Theorem}[section]
\newtheorem{lemma}[theorem]{Lemma}
\theoremstyle{definition}
\newtheorem{definition}[theorem]{Definition}
\newtheorem{assumption}[theorem]{Assumption}
\theoremstyle{remark}
\newtheorem{remark}[theorem]{Remark}

\newcommand{\algoname}{accumulative sub-sampling}

\graphicspath{ {./Image/} }
\usepackage{url}

\begin{document}

\newcommand{\mytitle}{
Accumulation of Sub-Sampling Matrices with Applications to Statistical Computation
}

\title{\mytitle
\thanks{This work is an extension of \cite{chen2021accumulations}.
}}

\author{
\centerline{
\name Yifan Chen$^{\dagger}$~~~
\name Yun Yang$^{\ddagger}$
} \\[1ex]
\centerline{$^\dagger$ \addr Departments of Mathematics and Computer Science, Hong Kong Baptist University}
\centerline{$^\ddagger$ \addr Department of Mathematics, University of Maryland, College Park}
}

\maketitle

\begin{abstract}%
With appropriately chosen sampling probabilities, sampling-based random projection can be used to implement large-scale statistical methods, substantially reducing computational cost while maintaining low statistical error. However, computing optimal sampling probabilities is often itself expensive, and in practice one typically resorts to suboptimal schemes. This generally leads to increased time and space costs, as more subsamples are required and the resulting projection matrices become larger, thereby making the inference procedure more computationally demanding. In this paper, we extend the framework of sampling-based random projection and propose a new projection method, \emph{\algoname{}}. By carefully accumulating multiple such projections, \algoname{} improves statistical efficiency while controlling the effective matrix size throughout the statistical computation.
On the theoretical side, we quantify
how the quality of the subsampling scheme affects the error in approximating matrix products and positive semidefinite matrices, and show how the proposed accumulation strategy mitigates this effect. Moreover, we apply our method to statistical models involving intensive matrix operations, such as eigendecomposition in spectral clustering and matrix inversion in kernel ridge regression, and demonstrate that reducing the effective matrix size leads to substantial computational savings. Numerical experiments across a range of problems further show that our approach consistently improves computational efficiency compared to existing random projection baselines under suboptimal sampling schemes.
\end{abstract}

\noindent%

\section{Introduction}

Implementation of most statistical models inevitably involves intense matrix computation due to ubiquitous operations on design matrices.
Such matrix operations are usually the computational bottleneck in statistical models, and a flurry of works~\citep{drineas2006fast,drineas2006sampling,woodruff2014sketching,yang2017randomized,chen2021accumulations} turn to \emph{random projection} for accelerating those matrix operations.
The core motivation behind this paradigm is
to reduce the dimension of the original matrix, which lowers the time and space complexity of downstream statistical computation.

Particularly, \emph{importance sampling} can be formulated as a \emph{data-adaptive} random projection approach, 
whose choice of sampling probabilities 
depend on input data.
Compared to data-oblivious projection methods (invariant with input data and attaining \emph{data-oblivious} universal high-probability approximation error guarantees), 
importance sampling is more efficient to perform\footnote{
To maintain the focus on our primary topic, sampling-based random projection, we defer the extended discussion of this comparison to Supplement~\ref{sec:compare-sparsity}.} 
and thus has been adopted in scaling up various modern data science applications,
such as layer-wise sampling for graph convolutional networks~\citep{chen2023calibrate} and attention approximation for large language models~\citep{chen-etal-2022-sketching}. 
In this paper, we revisit the underexplored impact of sampling quality on the required projection dimension, which directly determines the time and space complexity of subsequent statistical computations.
We further propose a new sampling-based projection method, \algoname{}, which controls the projection dimension even when suboptimal sampling schemes are employed.

In more technical terms, random projection can be formulated through a sketching matrix
$\mtx{\Pi} \in \mb R^{d \times n}$.
Given an $n \times p_A$ design matrix $\mtx{A}$, one computes the sketched matrix
$\mtx{\Pi}\mtx{A}$ as a lower-dimensional proxy for $\mtx{A}$.
In the case of importance sampling, this sketching operation can be represented by a
sub-sampling matrix $\mtx{\Pi}$ (see Definition~\ref{def:subsampling}),
whose rows are independently drawn from the set of $n$-dimensional canonical basis vectors
according to prescribed sampling probabilities. In many applications, the goal is to approximate the matrix product $\mtx{A}^\mathsf{T}\mtx{B}$, where $\mtx{A} \in \mb R^{n \times p_A}$ and
$\mtx{B} \in \mb R^{n \times p_B}$. This is achieved by sampling rows from $\mtx{A}$ and $\mtx{B}$ and forming the estimator
$\mtx{A}^\mathsf{T}\mtx{\Pi}^\mathsf{T}\mtx{\Pi}\mtx{B}$.
The quality of the approximation is measured by the error
$\matnorm{\mtx{A}^\mathsf{T}\mtx{\Pi}^\mathsf{T}\mtx{\Pi}\mtx{B} - \mtx{A}^\mathsf{T}\mtx{B}}_{\rm X}$
for a chosen matrix norm $\matnorm{\cdot}_{\rm X}$, with smaller values indicating a more accurate estimator. Classical results in random projection theory show that, under optimal sampling probabilities
$p_i^\ast$, the expected Frobenius norm error
$\Expect\big[\matnorm{\mtx{A}^\mathsf{T}\mtx{\Pi}^\mathsf{T}\mtx{\Pi}\mtx{B} - \mtx{A}^\mathsf{T}\mtx{B}}_{\rm F}^2\big]$
is minimized \citep{drineas2006fast}.
Since these optimal probabilities are typically unavailable in practice, one instead employs
suboptimal sampling probabilities $p_i$ satisfying
\[
p_i \ge \beta\, p_i^\ast, \quad \forall i \in [n],
\]
for some quality parameter $\beta \in (0,1]$, where both $\sum_i p_i = 1$ and $\sum_i p_i^\ast = 1$.
This formulation captures the practical limitation that sampling probabilities can only be
approximately optimized.

As expected, \cite{drineas2006fast} noted that the minimal projection dimension $d$ increases proportionally to $1/\beta$, under a fixed error tolerance.
In much of the existing theoretical literature, this factor is treated as a constant; however, $1/\beta$ can be enormous.
One common scenario is when sampling probabilities are estimated from an initial pilot sample, in which case the sampling quality may deteriorate over time.

More critically, the factor $1/\beta$ may itself increase with the matrix size $n$.
For example, under uniform sampling, a commonly adopted practical scheme for kernel ridge regression (KRR), the required projection dimension $d$ can be as large as $\widetilde{\m O}(n)$ \citep{yang2017randomized}, where $\widetilde{\m O}$ denotes $\m O$ up to polylogarithmic factors.
In this case, the overall computational cost becomes $\m O(n d^2)$.

This difficulty reflects a broader dilemma: obtaining the optimal sampling probabilities is often as computationally expensive as performing the original matrix operations~\citep{alaoui2015fast}.
As a result, in practice one typically has access only to crude approximations of the optimal probabilities due to limited computational budgets, or resorts to uniform sampling altogether.
Under such suboptimal sampling schemes, the factor $1/\beta$, the required projection dimension $d$, and consequently the computational complexity of fitting statistical models can all grow rapidly. While prior work has extensively studied the derivation of optimal sampling probabilities $p_i^\ast$ in various settings, the problem of efficient statistical computation given already suboptimal sampling probabilities has received little attention.

\smallskip

\noindent \textbf{Contribution 1: A new methodology that significantly reduces the size of $\mtx{\Pi}\mtx{A}$ under suboptimal sub-sampling schemes, together with connections to established random projection methods.
}
In this work, we introduce a new sampling-based random projection framework, \algoname{}, which achieves a small projection dimension without requiring full knowledge of the optimal sub-sampling scheme.
Specifically, we construct the projection matrix as an accumulation of $m$ randomly signed sub-sampling matrices (see Definition~\ref{def:subsampling}).
This framework includes classical sub-sampling methods and Gaussian sketching (cf.\ Section~\ref{sec:related}) as special cases, obtained by setting $m=1$ and $m=\infty$, respectively.
By introducing the accumulation parameter $m$, the total number of sampled rows, $m d$, is allowed to exceed the projection dimension $d$.
This separation makes it possible to improve statistical accuracy by using more sub-samples while keeping $d$ small, thereby reducing the complexity of subsequent matrix operations.

In addition, we connect \algoname{} to the compositional sketching technique \citep[Remark~3]{DBLP:conf/icalp/CohenNW16}.
That work showed a composition of the form $\mtx \Pi = \mtx \Pi_1 \mtx \Pi_2$ can achieve both a favorable sketch dimension and fast matrix multiplication, when $\mtx \Pi_1$ is a computationally heavier but more accurate sketching method.
We show that \algoname{} can also be expressed in a compositional sketching form (see Section~\ref{sec:sketching_composition}), where $\mtx \Pi_1$ is a block-wise sketching matrix and $\mtx \Pi_2$ is a sub-sampling sketching matrix.
In particular, \cite{DBLP:conf/icalp/CohenNW16} suggested choosing $\mtx \Pi_1$ as a $d \times m d$ Gaussian sketching matrix; combined with the same $m d \times n$ sub-sampling matrix as $\mtx \Pi_2$, their approach incurs a computational cost of $\m O(p_A m d^2)$ to compute $\mtx{\Pi A}$, which exceeds the $\m O(p_A m d)$ complexity achieved by \algoname{}.

While structurally similar to \cite{DBLP:conf/icalp/CohenNW16}, the sketching matrix $\mtx \Pi_1$ in our method
alone does not provide accurate matrix approximation guarantees, 
which violates their requirement that the ``slow'' sketching matrix should be sufficiently powerful and prevents the direct application of existing compositional guarantees.
We therefore develop a tailored analysis of the cross terms incurred by \algoname{} in Section~\ref{sec:res}.
Our theoretical results show that the proposed projection method admits a projection dimension $d$ that depends only on the stable ranks\footnote{The stable rank of a matrix $\mtx A$ is defined as $\matnorm{\mtx A}_{\rm F}^2 / \matnorm{\mtx A}^2$, namely the squared ratio of its Frobenius norm to its spectral norm.} of the target matrices $\mtx{A}$ and $\mtx{B}$, while the threshold for $md$ depends on the quality of the sub-sampling probability estimates.
In particular, the dimension $d$ in our framework removes any dependence on the quality parameter $\beta$ and can be chosen as small as that required by Gaussian sketching (modulo poly-logarithmic factors), while still retaining the computational efficiency of sub-sampling-based methods.
We prove that in order to achieve the high probability guarantee
$\Prob{\matnorm{\mtx{A}^\mathsf{T} \mtx{\Pi}^\mathsf{T} \mtx{\Pi} \mtx{B} - \mtx{A}^\mathsf{T} \mtx{B}} > \varepsilon \matnorm{\mtx{A}} \matnorm{\mtx{B}}} < \rho$
(see Definition~\ref{def:amm} for details), it suffices for $d$ and $md$
to scale as $\wt{\m O}(s / \varepsilon)$ and $\wt{\m O}(s / (\beta \varepsilon))$, respectively.
In contrast, the naive sub-sampling procedure based on uniform sampling may require a projection dimension $d$ that is orders of magnitude larger.

\smallskip

\noindent \textbf{Contribution 2: \textit{Application} of \algoname{} to accelerate the computation of several statistical problems, with accompanying error analysis.
}
For data science practitioners, we illustrate the application of our methods to scaling up statistical computation involving intensive matrix multiplication, with accompanying error analysis.
We 
apply \algoname{} to accelerate several computationally expensive data science tasks,
including eigendecomposition in relevant statistical models and computations in kernel ridge regression (KRR; \citealt{williams2001using, yang2017randomized}).
We observe that our proposed method is well suited to these scenarios.

Specifically, we provide performance guarantees for our proposed method in these applications.
\emph{First}, a range of statistical methods~\citep{jolliffe1982note,DBLP:conf/nips/NgJW01,xu2017generalized} rely on singular value decomposition (SVD, or equivalently eigendecomposition for positive semidefinite matrices) to construct features.
In particular, for large graphs with $n$ nodes, one must handle an $n \times n$ adjacency matrix, for which the time complexity of a full SVD is $\m O(n^3)$.
A common approach is to use randomized SVD~\citep{halko2011finding} to efficiently compute a subset of singular vectors with cost $\m O(n d^2)$.
However, under low-quality sampling schemes, the projection dimension $d$ may need to be as large as $\wt{\m O}(n)$ to guarantee low approximation error.
\emph{Second}, we apply the proposed method to a representative computationally intensive nonparametric procedure, kernel ridge regression.
The key step in accelerating KRR is to obtain a rank-$d$ approximation of the $n \times n$ empirical kernel matrix $\mtx K$, which reduces the overall computational complexity to $\m O(n d^2)$.
Yet, under low-quality sampling schemes, the required projection dimension $d$ can again approach $\wt{\m O}(n)$, causing the computational cost of KRR to grow prohibitively large.
We show that our method can maintain a small projection dimension $d$ and enable fast subsequent matrix operations, even when optimal probabilities are unavailable.

\subsection{Related work}
\label{sec:related}

We revisit two representative random projection paradigms, namely sub-Gaussian maps and importance sampling (our focus), to provide clearer context.
These two paradigms have different emphases and have traditionally been studied independently.
Our goal in this review is to streamline the discussion of how they can be unified under the \algoname{} framework in Section~\ref{sec:def}.

In general, the paradigm of sub-Gaussian maps was developed to implement the \emph{Johnson--Lindenstrauss} (JL) \emph{transform}~\citep{johnson1984extensions},
a foundational tool in dimension reduction theory.
The JL transform is characterized by a sketching matrix $\mtx{\Pi} \in \mb R^{d\times n}$ such that, for any vector $\mtx z \in \mb R^n$ with $\|\mtx z\| = 1$, $\|\mtx{\Pi} \mtx z\|$ can concentrate around $1$ as well.
The most straightforward construction sets the entries of $\mtx{\Pi}$ to be independent and identically distributed (i.i.d.) sub-Gaussian random variables, with representative examples including dense sub-Gaussian maps \citep{vershynin2010introduction, DBLP:journals/siamrev/HalkoMT11} and very sparse (VS) random projections \citep[entries are zero with a certain probability;][]{achlioptas2001database, li2006very, ahmad2022p}.
To further accelerate the computation of $\mtx{\Pi}\mtx z$, one can employ structured constructions such as Fast JL transforms~\citep{ailon2009fast}, including the sub-sampled randomized Hadamard transform (SRHT)~\citep{sarlos2006improved, ailon2006approximate, lu2013faster, yang2017randomized},
or sparse JL transforms~\citep{dasgupta2010sparse, kane2014sparser}, such as Count Sketch~\citep{charikar2004finding}.

In addition to the data-oblivious methods, \textit{data-adaptive} projection methods are more widely adopted in data science.
These methods are mostly sampling-based,
and construct sketches by selecting a subset of design points according to probabilities $\{p_i\}_{i=1}^n$.
To obtain theoretical guarantees that adapt to input data, \cite{drineas2006fast} proposed a \emph{data-adaptive} sampling scheme that minimizes the Frobenius norm error
$\matnorm{\mtx{A}^\mathsf{T} \mtx{\Pi}^\mathsf{T} \mtx{\Pi} \mtx{B} - \mtx{A}^\mathsf{T} \mtx B}_{\rm F}$,
by using appropriate sub-sampling probabilities.
Beyond the original methodology, sampling-based random projection techniques have been extended to a range of applications, including least squares regression~\citep{drineas2006sampling} and graph sparsification~\citep{spielman2011graph}.
\cite{derezinski2021sparse} utilized leverage score sampling to preserve the inverse $(\mtx{A}^\mathsf{T} \mtx A)^{-1}$, for a tall matrix $\mtx A$ (where the dimension $p$ of $\mtx A$ must be a small constant).

Although all are naturally suited to dimension reduction, different projection paradigms exhibit distinct strengths.
For example, dense sub-Gaussian maps offer strong statistical accuracy but incur high computational cost, largely due to the use of dense sketching matrices~\citep{DBLP:conf/icalp/CohenNW16}.
Very sparse (VS) random projection methods, in contrast, are faster because they employ sparse sketching matrices; however, such sparse matrix operations are often inefficient on modern hardware, such as GPUs~\citep{dao2022monarch}.

\subsection{Paper outline}

The rest of this paper is organized as follows.
In Section~\ref{sec:bg}, we introduce preliminaries and notations to facilitate the discussion in subsequent sections.
Our proposed sketching technique is formally presented in Section~\ref{sec:approach}, where we also outline the key ideas underlying the proof.
Section~\ref{sec:res} contains the main theoretical results characterizing the properties of the proposed sketching matrices.
In Section~\ref{sec:application}, we investigate the theoretical implications of applying our methods to several learning tasks and compare their computational complexity with that of existing sketching methods under these settings.
Furthermore, in Section~\ref{sec:exp}, we demonstrate through numerical experiments on various statistical models that the empirical results are consistent with the theoretical guarantees established earlier.

\section{Background and Notations}
\label{sec:bg}

To ease our presentation, we first define some matrix notations heavily used along this paper.
Without further clarification we denote the $i$-th row of a $d \times n$ matrix $\mtx{\Pi}$ ($i \in [d]$) by $\mtx{\Pi}_{(i)}$ or $(\mtx{\Pi})_{(i)}$ (which is by convention a column vector), 
use $\mtx{\Pi}^{(j)}$ or $(\mtx{\Pi})^{(j)}$ to denote the $j$-th column of $\mtx{\Pi}$ ($j \in [n]$),
use $\mtx{\Pi}_k$ to represent the $k$-th matrix in the sequence $\{\mtx{\Pi}_k\}_{k=1}^m$,
let $(\mtx{\Pi}_{k})_{(i)}$ or the combined notation $\mtx{\Pi}_{k, (i)}$ be the $i$-th row of the $k$-th matrix in the sequence,
and use $\mtx{\Pi}_{ij}$ to represent the entry located at the $i$-th row and the $j$-th column of $\mtx{\Pi}$.

\noindent \textbf{(Randomly signed) Sub-sampling matrix}.
We define (randomly signed) sub-sampling matrices with sampling probabilities $\{p_j\}_{j=1}^n$ as follows.
\begin{definition}[Randomly signed sub-sampling matrix]
\label{def:subsampling}
Consider a random matrix $\mtx{\Pi} \in \mb R^{d \times n}$ with i.i.d.~rows. 
Each of its row $\mtx{\Pi}_{(i)}$ is randomly instantiated as
${1}/{\sqrt{d p_j}} \cdot \mtx e_j$ with probability $p_j, \forall j \in [n]$;
here, $\mtx e_j$ is the $j$-th column of the $n \times n$ identity matrix $\mtx{I}_n$. 
Then, $\mtx{\Pi}$ is called a sub-sampling matrix with probabilities~$\{p_j\}_{j=1}^n$.
Specifically, with $d$ i.i.d.\ standard Rademacher variables and the induced $d \times d$ random diagonal matrix $\mtx{R}_d$, 
we call the random matrix $\mtx{R}_d \mtx{\Pi}$ a randomly signed sub-sampling matrix with probabilities $\{p_j\}_{j=1}^n$.
\end{definition}
Randomly signed sub-sampling matrices are important components in \algoname{} (see Definition~\ref{def:accu_sketching}). 
It allows the accumulated rows $\sum \mtx{\Pi}_{(i)}$ to be re-scaled isotropic random vectors \citep[Definition~5.19]{vershynin2010introduction},
which means that $\Expect\brkt{\sum \mtx{\Pi}_{(i)} (\sum \mtx{\Pi}_{(i)})^\mathsf{T}}$ is a multiple of $\mtx I_n$. 
This property is desired for a general projection matrix $\mtx{\Pi}$ since we expect $\mtx{\Pi}^\mathsf{T} \mtx{\Pi}$ to be an unbiased estimator of $\mtx{I}_n$ and $\Expect\brkt{\mtx{A}^\mathsf{T} \mtx{\Pi}^\mathsf{T} \mtx{\Pi} \mtx{B}} = \mtx{A}^\mathsf{T} \mtx{B}$.

\noindent \textbf{Approximate matrix multiplication (AMM) property}.
The primary focus of this paper is to prove the performance guarantee of our method and utilize it to analyze the downstream applications using our proposed projection matrix $\mtx{\Pi}$.
Specifically, we hope the spectral (operator) norm of $\mtx{A}^\mathsf{T} \mtx{\Pi}^\mathsf{T} \mtx{\Pi} \mtx{B} - \mtx{A}^\mathsf{T} \mtx{B}$ to be small relative to $\matnorm{\mtx{A}} \matnorm{\mtx{B}}$.
The exact definition of the AMM property may vary depending on the fields of the works;
in this paper we stick to a common definition as follows, which shortly serves as a crucial tool for deriving performance guarantees in statistical computation.
\begin{definition}[Spectral norm guarantee for AMM]
\label{def:amm}
For two matrices $\mtx{A} \in \mb R^{n\times p_A}$, $\mtx{B} \in \mb R^{n\times p_B}$, and two constants $\varepsilon > 0$, $\rho < \frac12$, a random projection matrix $\mtx{\Pi} \in \mb R^{d \times n}$ is called to satisfy $(\varepsilon, \rho)$-AMM property for $\mtx A, \mtx B$, if
\begin{align}
\Prob{\matnorm{\mtx{A}^\mathsf{T} \mtx{\Pi}^\mathsf{T} \mtx{\Pi} \mtx{B} - \mtx{A}^\mathsf{T} \mtx{B}} > \varepsilon \matnorm{\mtx{A}} \matnorm{\mtx{B}}} < \rho.
\end{align}
Specifically, if $\mtx A = \mtx B$, $\mtx{\Pi}$ is said to satisfy $(\varepsilon, \rho)$-AMM property for $\mtx A$.
\end{definition}

\noindent \textbf{Stable rank}. 
The concept of matrix stable rank~\citep{DBLP:conf/icalp/CohenNW16} helps describe the minimal rank of the approximation $\mtx{A}^\mathsf{T} \mtx{\Pi}^\mathsf{T} \mtx{\Pi} \mtx{B}$ (or equivalently the smallest dimension of the projection matrix $\mtx{\Pi}$) required for accurately approximating $\mtx{A}^\mathsf{T} \mtx{B}$ in terms of the AMM property.
The stable rank $s$ of a matrix $\mtx{A}$ is defined as 
$s \defeq \matnorm{\mtx{A}}_{\rm F}^2 / \matnorm{\mtx{A}}^2$. 
It has been shown that any low-rank approximation of $\mtx{A}$ should have a rank of at least $s$ in order to accurately approximate $\mtx{A}$~\citep{DBLP:conf/icalp/CohenNW16}. 
In particular, if $\mtx{A}$ is orthonormal, the stable rank $s$ equals the rank of $\mtx{A}$ since all the components in $\text{span}(\mtx{A})$ are important.

Some common matrices in data science applications, such as empirical kernel matrices, have well-understood singular value decay rates in the literature. 
Consequently, it is possible to derive (the relative order of) stable ranks without assuming its growth (relative to sample size), as shown in Theorems~\ref{thm:tilde_K} and \ref{thm:k-satis}.

\noindent \textbf{Assumptions on the matrix sizes and the stable ranks.} 
Finally, as the main motivation behind sub-sampling-based sketching methods is to reduce computational complexity while maintaining numerical precision, an additional mild assumption is imposed to make the random projection computationally meaningful.
\begin{assumption}
\label{assum:size}
For the two matrices $\mtx{A} \in \mb R^{n\times p_A}, \mtx{B} \in \mb R^{n\times p_B}$, their dimensions $p_A, p_B$ are upper bounded by a polynomial of $n$, so that both $\log p_A$ and $\log p_B$ are $\m O(\log n)$. 
Let $s_A$ and $s_B$ be their stable ranks. We have $s:\, = \max(s_A, s_B) = o(n)$.
\end{assumption}
{Previous works usually take the dimensions $p_A, p_B$ as bounded constants, and in the rates/complexity thereof the leading factors thus implicitly depend on $p_A, p_B$ while not reflected.}
However, $p_A, p_B$ grow with the sample size $n$ in some data science applications, such as kernel ridge regression~\citep{alaoui2015fast}, spectral clustering~\citep{DBLP:conf/nips/NgJW01}, and attention approximation for language models~\citep{chen-etal-2022-sketching}.
We therefore generally assume their dimensions $p_A, p_B$ are upper bounded by a polynomial of $n$ in Assumption~\ref{assum:size} and carefully discuss the impacts of large $p_A, p_B$ along this paper.

The magnitude of the stable rank $s$ decides the intrinsic difficulty of approximating $\mtx{A}^\mathsf{T}\mtx{B}$, 
and one can informally take $\m O(p_A p_B s)$ as the complexity lower bound for any effective approximation $\mtx{A}^\mathsf{T} \mtx{\Pi}^\mathsf{T} \mtx{\Pi} \mtx{B}$~\citep{DBLP:conf/icalp/CohenNW16}.
If Assumption~\ref{assum:size} is not satisfied, then there will be no significant improvement in computational complexity over the original complexity of $\mathcal{O}(p_A p_B n)$ to compute $\mtx{A}^\mathsf{T} \mtx{B}$. 
This explains why previous works on approximate matrix multiplication usually assume $p_A, p_B$, the upper bounds of $s$, are small constants.
In light of the considerations, for all the theoretical results presented below, Assumption~\ref{assum:size} is assumed to hold by default; otherwise, low-rank approximation methods does not benefit the computation.

\section{Accumulative Sub-Sampling}
\label{sec:approach}

Our proposed method is built upon the observation that sub-sampling matrices and Gaussian sketching matrices can be treated as two special cases under a unified accumulation framework (detailed below).
We first illustrate this connection between Gaussian sketching and sub-sampling and formulate our proposed method, which we call ``\algoname{},'' in Section~\ref{sec:def}. 
We then investigate the proposed projection method in the rest of this section.

\subsection{Connecting sub-sampling with Gaussian sketching}
\label{sec:def}

Assuming that $\mtx{\Pi}_k$'s are i.i.d. randomly signed sub-sampling matrices defined in Definition~\ref{def:subsampling}, we observe that $\mtx{\Pi} = \lim_{m\to\infty} \frac{1}{\sqrt{m}} \sum_{k=1}^m \mtx{\Pi}_k$ is exactly a Gaussian sketching matrix, considering the vector central limit theorem~\citep[Theorem 3.10.7]{durrett2019probability}.
By setting the number of summands $m$ as $1$ or $\infty$ in the formula $\frac{1}{\sqrt{m}} \sum_{k=1}^m \mtx{\Pi}_k$, we recover classical importance sampling and Gaussian sketching under our accumulation framework. 

This framework introduces an accumulation parameter $m$.
As shown in Theorem~\ref{thm:amm_asym}, both $d$ and $md$ solely need to exceed certain thresholds to attain desired approximation performance. 
The threshold mechanism partially explains why Gaussian sketching usually has better performance than sub-sampling sketching with the same projection dimension $d$, as in Gaussian sketching ($m=\infty$), $md$ is never the bottleneck.

However, our lower bound condition on $md$ implies that requiring $m=\infty$ in Gaussian sketching is computationally unnecessary, as it leads to a dense sketching matrix along with heavy computation. Instead, by reducing $m$ to a finite number, we can still achieve comparable embedding performance as long as $md$ exceeds a threshold depending on the quality of the importance sampling.
When we only have access to suboptimal sampling probabilities, \algoname{} can make reasonable use of the information within sampling probabilities while reducing projection dimension.
We define our proposed projection matrices as follows.
\begin{definition}[$(m, d, \{p_j\}_{j=1}^{n})$-\algoname{} matrix]
\label{def:accu_sketching}
Let the $d\times n$ matrices $\{\mtx{\Pi}_k\}_{k=1}^m$ be a sequence of i.i.d.~randomly signed sub-sampling matrices with probabilities $\{p_j\}_{j=1}^n$.
We call $\mtx{\Pi}$ an $(m, d, \{p_j\}_{j=1}^{n})$-\algoname{} matrix if it has the same distribution as $\frac{1}{\sqrt{m}}\sum_{k=1}^m \mtx{\Pi}_k$.
\end{definition}

We close the subsection with the comparison of \algoname{} and another sparse method, very sparse (VS) random projection. 
VS follows the paradigm of sub-Gaussian maps and specifies each entry of the random projection matrix as a Rademacher variable with probability $p$ or zero with probability $1-p$.
In light of these, the main differences between \algoname{} and VS are as follows:
first, our method is based on importance sampling and thus data-adaptive, while VS performs uniform sampling and is data-oblivious;
second, projection matrices in our method have a fixed number of non-zero elements in each row, which guarantees that the algorithm will not randomly run out of memory in practice.

\subsection{Overview of proof techniques}
\label{sec:proof_overview}

To provide theoretical guarantees of \algoname{}, we follow a standard proof scheme of applying a matrix Bernstein inequality to analyze sub-sampling sketching, while carefully leveraging the structure (i.e., the sparse pattern) of \algoname{} matrices. 

Concretely, under our new framework, we still express the discrepancy as a sum of independent random matrices, namely 
\begin{align*}
\mtx{A}^\mathsf{T} \mtx{\Pi}^\mathsf{T} \mtx{\Pi} \mtx{B} - \mtx{A}^\mathsf{T} \mtx{B} = \sum_{i=1}^d \mtx{A}^\mathsf{T} \Big( \mtx{\Pi}_{(i)} (\mtx{\Pi}_{(i)})^\mathsf{T} - \frac{1}{d} \mtx I \Big) \mtx{B}.
\end{align*}
This form naturally fits into the framework of random matrix concentration \citep{tropp2012user}. To apply a matrix Bernstein inequality, we need to properly bound the spectral norm of $\mtx X_i \defeq \mtx{A}^\mathsf{T} \big( \mtx{\Pi}_{(i)} (\mtx{\Pi}_{(i)})^\mathsf{T} - \frac{1}{d}I \big) \mtx{B}$ and $\Expect \brkt{\mtx{X}_i \mtx{X}_i^\mathsf{T}}, \Expect \brkt{\mtx{X}_i^\mathsf{T} \mtx{X}_i}$ for all $i \in [d]$. Overall, due to the averaging over $m$ independent sub-sampling matrices, \algoname{} can reduce the norm of each $\mtx X_i$ by an extra factor of $\sqrt{m}$ from regular sub-sampling sketching (i.e.,~$m=1$) while keeping $\matnorm{\Expect \brkt{\mtx{X}_i \mtx{X}_i^\mathsf{T}}}, \matnorm{\Expect \brkt{\mtx{X}_i^\mathsf{T} \mtx{X}_i}}$ controlled (these terms exhibit an improved dependence on the sampling quality $\beta$ with an appropriate $m$; see the proof of Theorem~\ref{thm:amm_asym} in Supplement~\ref{sec:amm_asym}), thereby attaining the same approximation accuracy as regular sub-sampling but with substantially reduced projection dimension.

Technically, the analysis for regular sub-sampling greatly benefits from the unique property of $\mtx{\Pi}_{(i)} \mtx{\Pi}_{(i)}^\mathsf{T}$, which is a diagonal matrix with only one non-zero element. In this case, $\matnorm{\mtx{X}_i}$'s are uniformly bounded by the maximal squared $\ell_2$ norm of each re-scaled row from $\mtx{A}, \mtx B$, and the number of non-zero terms in $\Expect \brkt{\mtx{X}_i \mtx{X}_i^\mathsf{T}}, \Expect \brkt{\mtx{X}_i^\mathsf{T} \mtx{X}_i}$ is significantly reduced due to the sparsity of $\mtx{\Pi}_{(i)} \mtx{\Pi}_{(i)}^\mathsf{T}$.
However, in \algoname{}, $\mtx{\Pi}_{(i)} \mtx{\Pi}_{(i)}^\mathsf{T}$ is generally no longer diagonal once $m > 1$. Therefore, we need to address the thorny issue: $\Expect \brkt{\mtx{X}_i \mtx{X}_i^\mathsf{T}}, \Expect \brkt{\mtx{X}_i^\mathsf{T} \mtx{X}_i}$ are more complex than in the sub-sampling case ($m=1$) due to the additional cross terms. Consequently, the na\"ive  bound for $\matnorm{\mtx{X}_i}$ can become loose because, in the worst case, the $m$ sub-samples in $\mtx{\Pi}_{(i)}$ are identical, magnifying the maximal $\ell_2$ norm by a factor of $\sqrt{m}$. This, in turn, deteriorates the high probability error bound.

To handle the lack of a diagonal structure in \algoname{}, we carefully analyze the cross terms in $\Expect \brkt{\mtx{X}_i \mtx{X}_i^\mathsf{T}}, \Expect \brkt{\mtx{X}_i^\mathsf{T} \mtx{X}_i}$ and utilize a vector Bernstein inequality (see Theorem~\ref{thm:intro-bernstein-vect} in the supplement) to obtain a high probability bound for $\matnorm{\mtx{X}_i}$.
Specifically, to deal with the cross terms, we divide all the non-zero terms into four groups and separately compute them. After combining the four intermediate results, the final bound can be considerably improved and simplified.
In addition, we bound $\matnorm{\mtx{X}_i}$ by $\|\mtx{A}^\mathsf{T} \mtx{\Pi}_{(i)}\| \cdot \|\mtx{B}^\mathsf{T} \mtx{\Pi}_{(i)}\|$, where $\mtx{A}^\mathsf{T} \mtx{\Pi}_{(i)}, \mtx{B}^\mathsf{T} \mtx{\Pi}_{(i)}$ are vectors. Once a high probability bound is obtained, a simple union bound leads to the uniform bound for $\matnorm{\mtx{X}_i}$'s.
Combining all the pieces, we use the standard matrix Bernstein inequality (see Theorem~\ref{thm:intro-bernstein}) to obtain the final bound on $\matnorm{\mtx{A}^\mathsf{T} \mtx{\Pi}^\mathsf{T} \mtx{\Pi} \mtx{B} - \mtx{A}^\mathsf{T} \mtx{B}}$.

\subsection{A compositional sketching perspective}
\label{sec:sketching_composition}

The purpose of accumulating $m$ sub-sampling matrices is to compress the dimension of the random sketch $\mtx{\Pi} \mtx{A}$ from $md$ to $d$. 
This objective is related to previous works~\citep{dasgupta2010sparse, DBLP:conf/icalp/CohenNW16}, which composed different data-oblivious sketching matrices $\mtx{\Pi}_1, \mtx{\Pi}_2$ to {obtain a small-size sketching matrix $\mtx \Pi = \mtx{\Pi}_1 \mtx{\Pi}_2$ and to facilitate fast matrix multiplication for $\mtx{\Pi A}$ without sacrificing much numerical accuracy.}
{For example, in the sparse JL transform}~\citep{dasgupta2010sparse},
the authors constructed a $d \times mn$ Count Sketch matrix $\mtx \Pi_1$ and a fixed $mn \times n$ sparse ``accumulation matrix'' $\mtx \Pi_2$ (assuming $m=2, n=3$):
\begin{align*}
\mtx{\Pi}_2 = \left[\begin{array}{c c c} 
 1 & 0 & 0 \\ 
 1 & 0 & 0 \\ 
 0 & 1 & 0 \\ 
 0 & 1 & 0 \\ 
 0 & 0 & 1 \\ 
 0 & 0 & 1
\end{array}\right],
\end{align*}
which accumulates $\mtx \Pi_1$ at the cost of a high intermediate project dimension $mn$ after applying $\mtx \Pi_2$ to the target matrix $\mtx A$.
The consequent composition $\mtx{\Pi}_1 \mtx{\Pi}_2$ is proved to be a qualified JL transform with proper $d$ and $m$.

To reduce the row dimension $mn$ of $\mtx{\Pi}_1^\mathsf{T}$ and/or $\mtx{\Pi}_2$ from much higher than $n$ to a smaller number, \cite{DBLP:conf/icalp/CohenNW16} suggested that a composition of a small and dense ``slow'' sketching matrix $\mtx{\Pi}_1 \in \mb R^{d \times md}$ with $d\ll n$ and a ``fast'' and relatively large sketching matrix $\mtx{\Pi}_2 \in \mb R^{md \times n}$ can improve the computational efficiency by reducing the final sketching dimension while maintaining the desired approximation accuracy. 
Particularly, $\mtx{\Pi}_1$ is chosen as a sub-Gaussian map, and $\mtx{\Pi}_2$ is suggested to be SRHT, whose matrix multiplication with any vector $\mtx a$ is fast.

Interestingly, our proposed \algoname{} matrix can also be rewritten in a composition form, which extends the idea from data-oblivious sketching to data-adaptive sketching.
{For an \algoname{} matrix $\mtx \Pi = \mtx{\Pi_1 \Pi_2}$}, the ``fast'' matrix $\mtx{\Pi}_2 \in \mb R^{md \times n}$ is a radomly signed sub-sampling matrix;
the ``slow'' matrix $\mtx{\Pi}_1 \in \mb R^{d \times md}$ is accordingly constructed as $\paren{\mtx{\Pi}_{1, (1)}, \dots, \mtx{\Pi}_{1, (d)}}^\mathsf{T}$,
where $\mtx{\Pi}_{1, (i)}$ is a sparse vector in which only its $(m(i-1)+1)$-th to $(mi)$-th elements are i.i.d.\ (re-scaled) Radematcher variables, $\forall i \in [d]$.
An illustrative example of a specific accumulation matrix $\mtx{\Pi}_1$ with $m=2, d=3$ is given as 
\begin{align}
\label{eqn:accu-mtx}
\mtx{\Pi}_1 = \left[\begin{array}{c c c c c c} 
\pm 1 / \sqrt{2} & \pm 1 / \sqrt{2} & 0 & 0 & 0 & 0 \\ 
0 & 0 & \pm 1 / \sqrt{2} & \pm 1 / \sqrt{2} & 0 & 0 \\ 
0 & 0 & 0 & 0 & \pm 1 / \sqrt{2} & \pm 1 / \sqrt{2}
\end{array}\right].
\end{align}
While this compositional sketching perspective simplifies the implementation of our sketching method, it cannot facilitate the theoretical analysis of our method. 
In particular, a key proof step in \cite{DBLP:conf/icalp/CohenNW16} relies on the use of the triangle inequality so that $\matnorm{(\mtx{\Pi A})^\mathsf{T}(\mtx{\Pi B})-\mtx{A} ^\mathsf{T} \mtx B}$ can be upper bounded by
\begin{equation}
\begin{split}
\underbrace{\matnorm{\left(\mtx{\Pi}_{1} \mtx \Pi_{2} \mtx A\right)^\mathsf{T} \left(\mtx{\Pi}_{1} \mtx \Pi_{2} \mtx{B}\right)
- \left(\mtx{\Pi}_{2} \mtx{A}\right)^\mathsf{T} \left(\mtx{\Pi}_{2} \mtx{B}\right)}}_\mathrm{(\Rom 1)}\ \ +  \ \ \matnorm{\left(\mtx{\Pi}_{2} \mtx{A}\right)^\mathsf{T} \left(\mtx{\Pi}_{2} \mtx{B} \right) - \mtx{A}^\mathsf{T} \mtx{B}},
\end{split}
 \label{eqn:composition}
\end{equation}
whose success replies on the property that $\mtx{\Pi}_1$ should effectively sketch both $\mtx{\Pi}_{2} \mtx{A}$ and $\mtx{\Pi}_{2} \mtx{B}$ in order to control term $\mathrm{(\Rom 1)}$ \citep{DBLP:conf/icalp/CohenNW16}. 
However, in our method, the sketching matrix $\mtx{\Pi}_1$ is an ``unqualified'' sketching matrix which violates their assumptions for directly applying the matrix Bernstein inequality to bound term $\mathrm{(\Rom 1)}$. 

Specifically, in our method, although $\mtx{\Pi}_1^\mathsf{T} \mtx{\Pi}_1$ remains an unbiased estimator of the $md$-by-$md$ identity matrix $\mtx{I}_{md}$, the $i$-th $m$-by-$m$ rank-$m$ diagonal block in $\mtx{I}_{md}$ is approximated by solely a single rank-one matrix $\mtx{\Pi}_{1, (i)} \mtx{\Pi}_{1, (i)}^\mathsf{T}$. 
Therefore, we cannot expect the block-wise matrix $\mtx{\Pi}_1$ alone to perform well on approximation. 
Nonetheless, our theoretical and empirical findings show that although individually, a block-wise sketching matrix cannot achieve high accuracy, combining it with a sub-sampling matrix can achieve an effective trade-off between accuracy and efficiency.

\smallskip
\noindent \textbf{A new baseline}. This compositional perspective of our \algoname{} motivates \emph{a variant} which takes $\mtx{\Pi}_1$ as a sub-Gaussian sketching matrix. 
Since a sub-Gaussian matrix is a qualified data-oblivious projection matrix, the aforementioned proof technique based on decomposition~\eqref{eqn:composition} in \cite{DBLP:conf/icalp/CohenNW16} is applicable. 
The resulting upper bound from their analysis can save a $\log n$ factor compared to our analysis; see Appendix~A.3 in \cite{DBLP:conf/icalp/CohenNW16}). 
However, the time complexity of this sub-Gaussian sketching variant is $\m O(nmd^2)$, which exceeds not only the $\m O(nmd)$ complexity of \algoname{}, but also the typical complexity $\m O(n d^2)$ of downstream tasks (e.g., randomized SVD and kernel ridge regression, see Sections~\ref{sec:randomized-SVD} and~\ref{sec:krr}). 
Moreover, \algoname{} has an advantage over the sub-Gaussian variant in that it can preserve the sparsity of $\mtx{A, B}$ since both $\mtx{\Pi}_1$ and $\mtx{\Pi}_2$ are also sparse; 
this property is useful when addressing sparse objects such as graph Laplacian matrices, as a dense intermediate matrix $\mtx{\Pi A}$ takes much memory for large-scale statistical models. 
In addition to the analysis above, we provide extensive empirical comparisons between \algoname{} and its variants in Section~\ref{sec:exp}.

\subsection{Characteristics of \algoname{} in approximating matrix product}
\label{sec:exp_amm_pre}

We first evaluate and compare our \algoname{} under multiple configurations for approximating matrix product via a toy example in this subsection. 
More comprehensive numerical experiments on AMM and other downstream tasks are provided in Section~\ref{sec:exp}.  
Specifically, we compare \algoname{} (with accumulation $m=8$) to two sketching methods in the extreme cases: Gaussian sketching ($m=\infty$) and sub-sampling sketching ($m=1$).

\begin{figure}[t]
\centering
\includegraphics[width=0.85\textwidth]{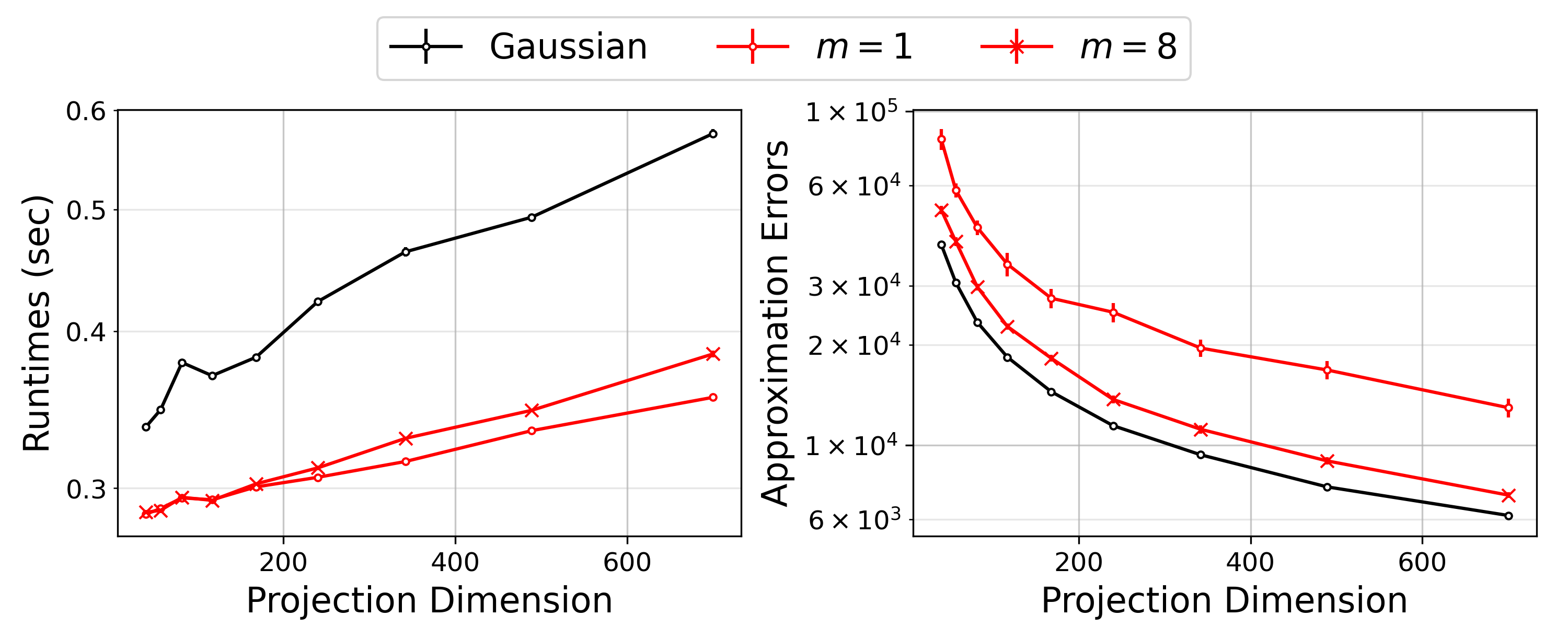}
\caption{\label{Fig:AMM_pre} \textbf{Runtime and approximation error for approximating matrix product}. 
The left panel shows runtime versus projection dimension $d$, and the right panel shows approximation error versus $d$.
Regular sub-sampling ($m=1$, the red curves with circle markers) with uniform importance (large $1/\beta$) achieves the highest efficiency but yields the largest approximation error for a given projection dimension.
}
\vspace{-3ex}
\end{figure}
We constructed the matrices to be multiplied as follows: starting with a $1000 \times 1000$ matrix with i.i.d.\ standard Gaussian elements, we further left-multiply it by a $1000 \times 1000$ diagonal matrix with i.i.d.\ standard Gaussian diagonal elements to form both $\mtx A$ and $\mtx B$ (to make the matrices ``non-uniform''). To simplify the evaluation process, we followed the common practice and used uniform sub-sampling for all the related sketching methods. The results, shown in Figure~\ref{Fig:AMM_pre}, are averaged over 30 replicates, and associated standard errors are also indicated in the figures.

We characterized these methods from two aspects: approximation accuracy and computational efficiency. Examining Figure~\ref{Fig:AMM_pre} in more detail, we see that our method achieves the advantages of the two extremes in our framework. Its accuracy is similar to Gaussian sketching, while also enjoying the low runtime of the same order as the vanilla sub-sampling method ($m=1$) when the accumulation size is moderate ($m=8$). We attribute the remarkable efficiency of our method to the sparsity of the sketching matrix and the highly optimized matrix addition operations on hardware.

\section{Theoretical Results for Accumulative Sub-Sampling}
\label{sec:res}

We present several important properties of our method, which serve as the intermediate theoretical tools for investigating its application to statistical computation. 
Throughout the paper, we assume that we can only access an \emph{approximation} to the optimal importance sampling probabilities $\{p^*_j\}_{j=1}^n$ for a certain task, where $n$ is the number of rows in the matrices $\mtx{A}$ and $\mtx{B}$. 
We impose the condition that the employed sampling probabilities satisfy $p_j \geq \beta \cdot p^*_j$ for all $j=1,\dots,n$, where $\beta \in (0, 1]$ is a parameter quantifying the closeness between $\{p_j\}_{j=1}^n$ and $\{p^*_j\}_{j=1}^n$. The threshold for $md$ mentioned above is later shown to be proportional to $1/\beta$. 
We also note that our condition includes the case of simple uniform sub-sampling probabilities $p_j = 1/n$ as a special case. 
However, in some extreme cases, the uniform probabilities may lead to a large lower bound threshold (close to $n$) for the required projection dimension $d$ in regular importance sampling.

Our first fundamental result is the following spectral norm guarantee of our method for approximating matrix product.
The complete proof is provided in Supplement~\ref{sec:amm_asym}, and we note, the letter $C$ in the bounds denotes a positive absolute constant whose value may change from line to line throughout this paper.
\begin{theorem}
\label{thm:amm_asym}
Let $\mtx{A} \in \mb R^{n\times p_A}, \mtx{B} \in \mb R^{n\times p_B}$ be matrices with stable ranks $s_A, s_B \leq s$, and let $\varepsilon > 0, \rho < 1/2$ be given constants. 
Suppose $\mtx{\Pi}$ is an $(m, d, \{p_j\}_{j=1}^{n})$-\algoname{} matrix, where its sampling probabilities $p_j$'s satisfy
\begin{align*}
p_j \geq \beta \frac{\max\{\|\mtx{A}_{(j)}\|^2 / \matnorm{\mtx{A}}^2, \|\mtx{B}_{(j)}\|^2 / \matnorm{\mtx{B}}^2\}}{\sum_{j'=1}^{n} \max\{\|\mtx{A}_{(j')}\|^2 / \matnorm{\mtx{A}}^2, \|\mtx{B}_{(j')}\|^2 / \matnorm{\mtx{B}}^2\}}, \quad \forall j = 1,\dots,n,
\end{align*}
for some $\beta \in (0, 1]$. 
There exists a positive absolute constant $C$ such that if
\begin{align*}
d \geq C \frac{s}{\varepsilon} \log\frac{n}{\rho} \max\left\{\frac{1}{\varepsilon}, \log \frac{n}{\rho}\right\}, \quad
md \geq C \frac{s}{\beta \varepsilon} \log\frac{n}{\rho} \max\left\{\frac{1}{\varepsilon}, \log^2 \frac{n}{\rho}\right\},
\end{align*}
then $\mtx{\Pi}$ satisfies $(\varepsilon, \rho)$-AMM property for $\mtx{A}$ and $\mtx{B}$ (see Definition~\ref{def:amm}).
\end{theorem}

The bound in the theorem above has several implications. Firstly, it verifies our previous claim that $d$ and $md$ only need to exceed certain thresholds. 
Moreover, the sampling quality coefficient $\beta$ plays an important role in the approximation. 
If we have precise knowledge of the optimal sub-sampling scheme ($\beta=1$) and require high accuracy ($\varepsilon$ small), regular sub-sampling ($m=1$) alone can provide efficient and accurate approximation. 
However, in most practical cases, obtaining the optimal sub-sampling scheme is intractable. 
In such cases, our method, along with a suboptimal sub-sampling scheme, can achieve a better balance: it can nearly approach the same accuracy as using $md$ sub-samples while maintaining the projection dimension as $d$.

\begin{remark}
\normalfont
To ease the analysis, we constructed an unintuitive sub-sampling scheme where $p_j$ is proportional to $\max\{\|\mtx{A}_{(j)}\|^2/\matnorm{\mtx{A}}^2, \|\mtx{B}_{(j)}\|^2/\matnorm{\mtx{B}}^2\}$ instead of $\|\mtx{A}_{(j)}\| \cdot \|\mtx{B}_{(j)}\|$; the latter is proved to minimize $\matnorm{\Expect \brkt{\mtx{A}^\mathsf{T} \mtx{\Pi}^\mathsf{T} \mtx{\Pi} \mtx{B} - \mtx{A}^\mathsf{T} \mtx{B}}}_F$ \citep{drineas2006fast}.
Similar tricks can be found in previous analyses of matrix Chernoff bounds \citep{ahlswede2002strong, drineas2006sampling, tropp2015introduction}.
We remark that the probabilities we use are close to the optimal ones above, and those two will coincide with each other when $\|\mtx{A}_{(j)}\| = \|\mtx{B}_{(j)}\|, \forall j \in [n]$.
It turns out that the final bounds in Theorem~\ref{thm:amm_asym} are of the same order as those in Theorem~\ref{thm:amm_sym},
where $\mtx{A} = \mtx{B}$ and canonical sub-sampling schemes (setting $p_j \propto \|\mtx{A}_{(j)}\| \|\mtx{B}_{(j)}\|$) are adopted.
\end{remark}

Theorem~\ref{thm:amm_asym} directly implies the following statement for the special case where $\mtx{A} = \mtx{B}$. 
We omit its proof---the only difference therein is that instead of Theorem~\ref{thm:intro-bernstein} (rectangular matrix Bernstein), a symmetric matrix Bernstein is used in the final step.
\begin{theorem}
\label{thm:amm_sym}
Let $\mtx{A} \in \mb R^{n\times p}$ be a matrix with stable rank $s$. 
Suppose $\mtx{\Pi}$ is an $(m, d, \{p_j\}_{j=1}^{n})$-\algoname{} matrix with $p_j \geq \beta \frac{\|\mtx{A}_{(j)}\|^2}{\matnorm{\mtx{A}}_{\rm F}^2}, \forall j \in [n]$, for some $\beta \in (0, 1]$. 
There exists a positive absolute constant $C$ such that if
\begin{align*}
d \geq C \frac{s}{\varepsilon} \log\frac{n}{\rho} \max\left\{\frac{1}{\varepsilon}, \log \frac{n}{\rho}\right\}, \quad
md \geq C \frac{s}{\beta \varepsilon} \log\frac{n}{\rho} \max\left\{\frac{1}{\varepsilon}, \log^2 \frac{n}{\rho}\right\}, 
\end{align*}
then with two constants $\varepsilon>0, \rho < 1/2$, $\mtx{\Pi}$ satisfies the $(\varepsilon, \rho)$-AMM property for $\mtx{A}$.
\end{theorem}

\smallskip
\noindent \textbf{Results on \nystrom approximation.}
In addition to approximating matrix product, we investigate another popular technique, \nystrom approximation \citep{alaoui2015fast}, developed for approximating a positive semidefinite (PSD) matrix $\mtx{K} \in \mb R^{n\times n}$. 
This method approximates $\mtx{K}$ by a low-rank PSD matrix $\wt{\mtx{K}} \defeq \mtx{K} \mtx{\Pi}^\mathsf{T} (\mtx{\Pi} \mtx{K} \mtx{\Pi}^\mathsf{T})^\dagger \mtx{\Pi} \mtx{K}$, 
where $\mtx{A}^\dagger$ denotes the Moore–Penrose pseudoinverse of the matrix $\mtx{A}$.
{We summarize the results of applying \algoname{} to \nystrom approximation in Theorem~\ref{thm:tilde_K}, and provide the proof in Supplement~\ref{app:tilde_K} for the sake of self-containment.}
\begin{theorem}[Adapted from \cite{musco2017recursive}]
\label{thm:tilde_K}
Consider a PSD matrix $\mtx{K} \in \mb R^{n\times n}$. 
For some positive $\lambda < \matnorm{\mtx{K}}$, we define 
the leverage scores $\ell_j \defeq [\mtx{K} (\mtx{K} + \lambda \mtx{I})^{-1}]_{jj}, \forall j \in [n]$, 
and the statistical dimension $d_{\rm stat} \defeq \sum \ell_j$.
Suppose $\mtx{\Pi}$ is an $(m, d, \{p_j\}_{j=1}^{n})$-\algoname{} matrix with $p_j \geq \beta \frac{\ell_j}{d_{\rm stat}}$, for some $\beta \in (0, 1]$ and all $j \in [n]$. 
There exists an absolute constant $C$ such that for a constant $\rho < \frac12$, if  
\begin{align*}
d \geq C d_{\rm stat} \log^2 \frac{n}{\rho}, \quad
md \geq C \frac{d_{\rm stat}}{\beta} \log^3 \frac{n}{\rho},
\end{align*}
then $\wt{\mtx{K}} \psdle \mtx{K} \psdle \wt{\mtx{K}} + \lambda \mtx{I}$ with probability $1-\rho$.
Here $\psdle$ denotes the Loewner ordering: $\mtx{A} \psdle \mtx{B}$ means $\mtx{B} - \mtx{A}$ is positive semidefinite.
\end{theorem}
Shortly in Section~\ref{sec:krr}, we will apply the \nystrom approximation result to kernel ridge regression \citep[KRR,][]{williams2001using, yang2017randomized}, a renowned nonparametric model.

\section{Applications to Statistical Computation and Associated Theoretical Consequences}
\label{sec:application}

In this section, we demonstrate how our proposal can handle suboptimal sampling probabilities in fast computation of
some common statistical applications. 
We will illustrate that the spectral norm guarantees and dimension bounds provided by Theorems~\ref{thm:amm_asym} and~\ref{thm:amm_sym} can be immediately applied to the error analysis of statistical computation relying on low-rank approximations.

\subsection{Accelerated eigendecomposition in statistical computation}
\label{sec:randomized-SVD}

Eigendecomposition is heavily utilized in statistical models, while its formidable computational cost hinders the wide application of those models in large-scale data science problems.
Randomized SVD \citep[Section~1.6]{DBLP:journals/siamrev/HalkoMT11} was 
accordingly developed to address this issue (considering SVD is equivalent to eigendecomposition for a PSD matrix).
We then devote this subsection to the discussion about accelerating randomized SVD via our proposed \algoname{}.

Specifically, 
if one is interested in only the top-$k$ singular values (usually $k \leq d \leq p$) of a target matrix $\mtx{A} \in \mathbb{R}^{n \times p_A}$, SVD can be performed on the reduced matrix $\mtx{A} \mtx{Q} = \wt{\mtx{U}} \wt{\mtx{\Sigma}} \wt{\mtx{V}}^\mathsf{T}$, where $\mtx{Q} \in \mathbb{R}^{p_A \times d}$ is obtained from the QR decomposition of the sketched matrix $(\mtx{\Pi} \mtx{A})^\mathsf{T}$ (we denote the factorization by $\mtx{\Pi} \mtx{A} = \mtx R^\mathsf{T} \mtx Q^\mathsf{T}$).
\cite{DBLP:journals/siamrev/HalkoMT11} proposed to approximate the SVD of $\mtx A$ (i.e., obtaining $\mtx{U}$, $\mtx{\Sigma}$, and $\mtx{V}$) by the low-rank counterparts $\wt{\mtx{U}}$, $\wt{\mtx{\Sigma}}$, and $\mtx{Q} \wt{\mtx{V}}$;
keeping the first $k$ singular values/vectors from $\wt{\mtx{U}}$, $\wt{\mtx{\Sigma}}$, and $\mtx{Q} \wt{\mtx{V}}$, we denote the corresponding top-$k$ approximation as $\wt{\mtx{A}}_k$. 
We denote by $\mtx{A}_k$ the exact $k$-truncated SVD approximation of $\mtx{A}$, 
which is the optimal rank-$k$ approximation (w.r.t.\ the spectral norm).

The matrix $\wt{\mtx{A}}_k$ obtained by randomized SVD is the \emph{best} possible rank-$k$ approximation (w.r.t.\ the spectral norm) to $\mtx{A}$ that lies within the row space of ${\mtx{\Pi}} \mtx{A}$~\cite[Theorem~4]{DBLP:conf/icalp/CohenNW16}. 
This property connects to generalized linear regression for vector-valued outputs, whose estimator is $\mtx A \mtx{X} \defeq \mtx A (\mtx{A}^\mathsf{T} \mtx{A})^\dagger \mtx{A}^\mathsf{T} \mtx{B}$.
This estimator $\mtx A \mtx{X}$ can be efficiently approximated by replacing $\mtx{A}$ and $\mtx{B}$ with their sketched approximations $\mtx{\Pi}\mtx{A}$ and $\mtx{\Pi} \mtx{B}$, respectively,
which reads $\mtx A \wt{\mtx{X}} = \mtx A (\mtx{A}^\mathsf{T} \mtx{\Pi}^\mathsf{T} \mtx{\Pi} \mtx{A})^\dagger \mtx{A}^\mathsf{T} \mtx{\Pi}^\mathsf{T} \mtx{\Pi} \mtx{B}$ and notably lies in the row space of ${\mtx{\Pi}} \mtx{B}$ as well.
Considering the optimality of $\wt{\mtx{A}}_k$, the approximation error of arbitrary generalized linear regression estimators can serve as the upper bound for the approximation error of $\wt{\mtx{A}}_k$.

In more detail, if we run generalized linear regression and project $\mtx{A}$ onto $\mtx{A}_k$ using ${\mtx{\Pi}}$,
then the resulting approximate regression estimator is given by
\begin{align*}
\mtx{A}_k \wt{\mtx{X}} \defeq \mtx{A}_k \left(\mtx{A}_k^\mathsf{T} {\mtx{\Pi}}^\mathsf{T} {\mtx{\Pi}} \mtx{A}_k \right)^\dagger 
\mtx{A}_k^\mathsf{T} {\mtx{\Pi}}^\mathsf{T} {\mtx{\Pi}} \mtx{A}, 
\end{align*}
which clearly lies within the row space of ${\mtx{\Pi}} \mtx{A}$ and we know the approximation error $\matnorm{\wt{\mtx{A}}_k - \mtx{A}}^2 
\leq \matnorm{\mtx{A}_k \wt{\mtx{X}} - \mtx{A}}^2$.
In specifying the sampling scheme in $\mtx \Pi$ to minimize the upper bound $\matnorm{\mtx{A}_k \wt{\mtx{X}} - \mtx{A}}^2$, we follow the results on generalized linear regression (c.f.\ Theorem~\ref{thm:gr} in the supplement),
denote the left singular vectors of $\mtx A_k$ as $\mtx U_k$ and $\bar{\mtx A}_k \defeq \mtx{A} - \mtx{A}_k$ the residual,
and finally formulate the sampling probabilities as
\begin{align}
\label{eqn:lvg-rsvd}
p_j \geq \beta \frac{\max\{\|[\mtx{U_k}]_{(j)}\|^2, \|[\bar{\mtx A}_k]_{(j)}\|^2 / \matnorm{\bar{\mtx A}_k}^2\}}{\sum_{j'=1}^{n} \max\{\|[\mtx{U_k}]_{(j')}\|^2, \|[\bar{\mtx A}_k]_{(j')}\|^2 / \matnorm{\bar{\mtx A}_k}^2\}}, \quad \forall j = 1,\dots,n.
\end{align}
We then obtain the following result (the complete proof is deferred to Supplement~\ref{sec:grproof}). 
\begin{theorem}
\label{thm:rsvd}
For $\mtx{A} \in \mb R^{n\times p_A}$, we let $s$ be the maximum of $k$ and the stable rank of the residual $\bar{\mtx A}_k = \mtx{A} - \mtx{A}_k$.
Suppose $d \geq k$ and ${\mtx{\Pi}} \in \mb R^{d \times n}$ is an $\paren{m, d, \{p_j\}_{j=1}^{n}}$-\algoname{} matrix where $\{p_j\}_{j=1}^{n}$ \textbf{satisfy equation~\eqref{eqn:lvg-rsvd}}.
For given constants $\varepsilon > 0, \rho < \frac12$, there exists a positive absolute constant $C$ such that if
\begin{align*}
d \geq C \frac{s}{\sqrt \varepsilon} \log\frac{n}{\rho} \max\left\{\frac{1}{\sqrt \varepsilon}, \log \frac{n}{\rho}\right\}, \quad
md \geq C \frac{s^2}{k \beta \sqrt \varepsilon} \log\frac{n}{\rho} \max\left\{\frac{1}{\sqrt \varepsilon}, \log^2 \frac{n}{\rho}\right\},
\end{align*}
then $\matnorm{\bar{\mtx A}_k}^2 \leq \matnorm{\wt{\mtx{A}}_k - \mtx{A}}^2 
\leq \matnorm{\mtx{A}_k \wt{\mtx{X}} - \mtx{A}}^2
\leq (1+\varepsilon) \matnorm{\bar{\mtx A}_k}^2$ with probability at least $1-\rho$.
Furthermore, if one use \textbf{uniform sampling} and $p_j = \frac1n, \forall j \in [n]$, then under the dimension condition
\begin{align*}
d \geq C \frac{s}{\sqrt \varepsilon} \log\frac{n}{\rho} \max\left\{\frac{1}{\sqrt \varepsilon}, \log \frac{n}{\rho}\right\}, \quad
md \geq C \frac{n s}{k \sqrt \varepsilon} \log\frac{n}{\rho} \max\left\{\frac{1}{\sqrt \varepsilon}, \log^2 \frac{n}{\rho}\right\},
\end{align*}
one can attain $\matnorm{\bar{\mtx A}_k}^2 \leq \matnorm{\wt{\mtx{A}}_k - \mtx{A}}^2 
\leq \matnorm{\mtx{A}_k \wt{\mtx{X}} - \mtx{A}}^2
\leq (1+\varepsilon) \matnorm{\bar{\mtx A}_k}^2$ with probability at least $1-\rho$.
\end{theorem}

\noindent \textbf{Complexity analysis.} 
We first indicate that the complexity of directly performing SVD is $\m O(n p_A^2)$, where $p_A$ can be $\m O(n)$ in some cases of interest (e.g., spectral clustering discussed in Remark~\ref{rmk:spec-clustering}).
For randomized SVD, one needs to first compute ${\mtx{\Pi}} \mtx{A} \in \mb R^{d \times p_A}$ (the complexity depends on the random projection method in use) and perform QR decomposition with time complexity $\m O(p_A d^2)$.
One then calculates the product $\mtx A \mtx Q \in \mb R^{n \times d}$ with time cost $\m O(n p_A d)$;
the last SVD step will take $\m O(n d^2)$ time.
In summary, the overall complexity will be $\m O(n p_A d + (n + p_A) d^2)$.

Notably, it is vital to reduce the projection dimension $d$ in the application of randomized SVD.
If we simply apply uniform sampling, the smallest possible total number of sub-samples can be as large as $\wt{\m O}(n)$, and the subsequent SVD will be even more expensive than directly performing SVD to $\mtx A$.
However, with \algoname{}, we can restrict the projection dimension $d$ to $\widetilde{\mathcal{O}}(s)$, and the complexity of the follow-up operations is accordingly reduced.
Additionally, we remark the time complexity of our method to compute ${\mtx{\Pi}} \mtx{A}$ is $\widetilde{\mathcal{O}}(n p_A)$, which is the same as other candidate efficient projection methods.
\begin{remark}[Adaptation to spectral clustering]
\label{rmk:spec-clustering}
\normalfont
A common data-driven method, \emph{spectral clustering}~\citep{von2007tutorial}, is built on the eigendecomposition of a graph Laplacian matrix $\mtx{L}$ and can benefit from the technique of randomized SVD \citep{halko2011finding}. However, spectral clustering relies on the eigenvectors corresponding to the \emph{smallest} $k$ eigenvalues, different from the largest eigenvalues targeted by randomized SVD. Therefore, some adaptations are required.
In particular, we can specify a sufficiently large constant $C$ such that all the eigenvalues of $\mtx{\tilde L} \defeq C \mtx{I} - \mtx{L}$ are positive. This adaptation ensures that the eigenvectors associated with the largest eigenvalues of $\mtx{\tilde{L}}$ are exactly the eigenvectors for the smallest eigenvalues of $\mtx{L}$. For a normalized Laplacian matrix, we can choose $C = 2$, which results in the signless normalized Laplacian matrix~\citep{cvetkovic2007signless}.
\end{remark}

\subsection{Fast kernel ridge regression}
\label{sec:krr}

In this subsection, we demonstrate the efficacy of \algoname{} to accelerate kernel ridge regression (KRR) with a reasonable sacrifice in numerical accuracy. 
Technically, we replace the empirical kernel matrix (to be defined below) with its sketched approximation to maintain statistical accuracy while reducing the $\mathcal{O}(n^3)$ computation complexity for matrix inversion required in KRR.
We refer readers to the original conference proceedings article \cite{chen2021accumulations} for more details.

In a standard regression model with $n$ samples, we use $\mtx Y=(Y_1, \ldots, Y_n)^\mathsf{T}$ to denote the observed response vector and $x_1,x_2\ldots,x_n\in \mathcal X$ are their associated predicting variables (or design points). 
The underlying statistical model is $Y_i=f^\ast(x_i)+w_i$, with $w_i$'s as i.i.d.\ noises, and $f^\ast$ the unknown regression function to estimate.
In kernel ridge regression, an in-sample prediction of $\big(f^\ast(x_1),\ldots,f^\ast(x_n)\big)^\mathsf{T}$ is given by 
\begin{align}
\label{eqn:krr-fn}
\hat f_n :\,=\big(\hat f_n(x_1),\ldots,\hat f_n(x_n)\big)^\mathsf{T}=\mtx{K} (\mtx{K} + \lambda \mtx{I}_n)^{-1} \mtx Y;
\end{align}
here, $\lambda>0$ is a regularization parameter, $\mtx{K}$ is the \emph{empirical kernel matrix} whose $(i, j)$-th element is $\mtx{K}_{ij} \defeq \m K(x_i, x_j)$, and $\m K(\cdot, \cdot)$ is the deployed kernel function.

KRR is a powerful nonparametric method that leads to minimax-optimal estimation of the regression function $f^\ast$ if it belongs to the reproducing kernel Hilbert space associated with kernel $\m K$. 
However, it suffers from extremely high $\mathcal{O}(n^3)$ computation cost due to the inversion of $(\mtx{K} + \lambda \mtx{I}_n)$. 
It is therefore of practical importance to reduce the computational cost, while maintaining a guarantee that the in-sample mean squared approximation error
\begin{align}
\label{eqn:in-sample}
\|\hat{f_S} - \hat{f_n}\|_n^2 \defeq n^{-1} \sum_{j=1}^n \big|\hat{f_S}(x_j) - \hat{f_n}(x_j)\big|^2
\end{align}
remains small with high probability.
Here, 
\begin{align}
\label{eqn:krr-fS}
\hat{f_S}:\,=\big(\hat{f_S}(x_1),\ldots,\hat{f_S}(x_n)\big)^\mathsf{T} = \wt{\mtx{K}} (\wt{\mtx{K}} + \lambda \mtx{I}_n)^{-1} \mtx Y
\end{align}
is the approximated prediction of $f^\ast$ at the design points $\{x_i\}_{i=1}^n$ via replacing $\mtx K$ with its sketched version $\wt{\mtx{K}} \defeq \mtx{K} \mtx{\Pi}^\mathsf{T} (\mtx{\Pi} \mtx{K} \mtx{\Pi}^\mathsf{T})^\dagger \mtx{\Pi} \mtx{K}$.
\cite{yang2017randomized} have carefully analyzed the performance of kernel sketched approximation to KRR, 
and proved that the in-sample approximation error $\|\hat{f_S} - \hat{f_n}\|_n^2$ is relatively small (compared to the statistical minimax rate) once the so-called \textbf{``$K$-satisfiability''} (defined below) holds. \cite{liu2018nonparametric} extend the analysis from estimation to hypothesis testing based on estimators of KRR.

We then devote this subsection to the proof of the ``$K$-satisfiability'' with \algoname{} (which is the core contribution in our preceding version \citep{chen2021accumulations}). 
We first introduce auxiliary notations to ease the illustration of ``$K$-satisfiability''. 
We denote the SVD of the re-scaled empirical kernel matrix $n^{-1} K$ as $\mtx{U} \mtx{\Sigma} \mtx{U}^\mathsf{T}$, 
where the diagonal elements in $\mtx{\Sigma}$ are its singular values $\sigma_1 \geq \sigma_2 \geq \cdots \geq \sigma_n \geq 0$.
We further define a cutoff size $d_{\delta} \defeq \max\{j: \sigma_j > \delta^2\}$ to denote the effective statistical dimension, where $\delta$ is typically chosen as the so-called critical radius $\delta_{\rm crit}$ \citep[depending on $n$,][]{yang2017randomized}, which reflects the statistical error/hardness of the problem (independent of the numerical error and any algorithmic tuning parameters).
We further use $\mtx{U}_1\in \mb R^{n \times d_{\delta}}$ (resp.\ $\mtx \Sigma_1$) to denote the first $d_{\delta}$ columns of $\mtx U$ (resp.\ the first $d_{\delta} \times d_\delta$ diagonal block of $\mtx \Sigma$), 
and $\mtx{U}_2 \in \mb R^{n \times (n-d_\delta)}$ (resp.\ $\mtx \Sigma_2$) the matrix composed of the rest columns (resp.\ diagonal block). 
With those notations, we define $K$-satisfiability as follows.
\begin{definition}[$K$-satisfiability]\label{def:k_satis}
For an empirical kernel matrix $\mtx{K}$, a projection matrix $\mtx{\Pi}$ is $K$-satisfiable (for ${\delta}$) if there exists a constant $c > 0$ such that,
\begin{align*}
\matnorm{\mtx{U}_1^\mathsf{T} \mtx{\Pi}^\mathsf{T} \mtx{\Pi} \mtx{U}_1 - \mtx I_{d_{\delta}}} \leq 1/2, \quad \mathrm{and} \quad
\matnorm{\mtx{\Pi} \mtx{U}_2 \mtx{\Sigma}_2^{1/2}} \leq c \,{\delta}.
\end{align*}
\end{definition}
{
As proved by \cite{yang2017randomized}, if a projection matrix $\mtx \Pi$ is $K$-satisfiable for $\delta_{\rm crit}$ and $\lambda\geq 2\delta_{\rm crit}^2$, the sketched estimator $\hat{f_S}$ constructed with $\mtx \Pi$ can attain a desirable in-sample estimation error of order $\mathcal O(\lambda+\delta_{\rm crit}^2)$.} 
In Theorem~\ref{thm:k-satis} below, we verify the $K$-satisfiability property for a generic $\delta$, which will be chosen as $\delta_{\rm crit}$ to achieve the optimal estimation accuracy.

To specify the conditions on $d$ and $m$ for \algoname{} to be $K$-satisfiable, we construct an auxiliary matrix $\mtx{\Psi}$ such that $\mtx{\Psi} \mtx{\Psi}^\mathsf{T} = \mtx{K} (\mtx{K} + n \delta^2 \mtx{I})^{-1}$ and apply Theorem~\ref{thm:amm_sym}, 
in which the corresponding regularization parameter is $\lambda^\ast = \Theta(\delta_{\rm crit}^2)$ and statistical dimension is $d_{\rm stat}:\, = \sum_{j=1}^n \frac{\sigma_j}{\sigma_j + \lambda^\ast}$.
A formal statement is provided below; a proof and involved assumptions on the eigenvalue decay rate of the empirical kernel matrix $\mtx K$ can be found in Supplement~\ref{app:k-satis} (which is technical and irrelevant to \algoname{}).
\begin{theorem}[Conditions on $d$ and $m$ for KRR~\citep{chen2021accumulations}]
\label{thm:k-satis}
Let the kernel function $\m K(\cdot, \cdot)$ satisfy Assumptions \ref{Assu:1} and \ref{Assu:2}. 
Let $\lambda \geq 2\delta^2$ be the regularization parameter and $\rho \in (0, 1/2)$. 
Assume the sub-sampling probabilities $p_j \geq \beta \frac{\|\mtx{\Psi}_{(j)}\|^2}{\matnorm{\mtx{\Psi}}_{\rm F}^2}$, for some $\beta \in (0, 1]$ and all $j \in [n]$. 
There exists a constant $C > 0$ such that if an $(m, d, \set{p_j}_{j=1}^n)$-\algoname{} matrix $\mtx \Pi \in \mb R^{d \times n}$ satisfies
\begin{equation}
\begin{aligned}
d \geq C d_\delta \log^2(\frac{n}{\rho}), \quad
md \geq C \frac{d_\delta}{\beta} \log^3(\frac{n}{\rho}),
\end{aligned}
\end{equation}
then with probability at least $1-\rho$, the matrix $\mtx \Pi$ is $K$-satisfiable for $\delta$.
\end{theorem}
For most common kernel function $\m K$, the cutoff size $d_{\delta_{\rm crit}}$ at the critical radius and the statistical dimension $d_{\rm stat}$ are of the same magnitude with high probability; see Supplement~\ref{app:d_stat_delta} for a proof of this statement under the assumptions in Theorem~\ref{thm:k-satis}. 
As a result, Theorem~\ref{thm:k-satis} provides the guidance of how a minimal projection dimension~$d$, with \algoname{}, should depend on the statistical dimension $d_{\rm stat}$ of the kernel ridge regression.

\smallskip
\noindent \textbf{Complexity analysis}. The overall runtime for fast KRR using a sampling-based method can be as low as linear in $n$ (up to logarithmic factors) since users do not need to construct the whole matrix $\mtx K$. 
In comparison, for other non-sampling data-oblivious methods, the time complexity is at least $\Omega(n^2)$ due to the explicit construction of $\mtx K$. Specifically, the complexity of fast KRR using our method is $\Theta(n m d)$, and $md$ therein can be $\m O(\beta^{-1}d_{\rm stat}\log^3 n)$, according to Theorem~\ref{thm:k-satis} and its following remark. 

This is a significant improvement over data-oblivious methods since $d_{\rm stat}$ is $\m O(\sqrt{n})$ for most kernels;
the runtime of our method $\Theta(n m d)$ is thus much smaller than the $\Omega(n^2)$ complexity in using data-oblivious methods.
For self-containedness, here we lay out the statistical dimension $d_{\rm stat}$ of a \matern kernel (with smoothness parameter $\nu$ being a positive half integer) as an example. 
Specifically, for a fixed input dimension $p_X \geq 1$, when the design $\{x_i\}_{i=1}^n \subset \m X \subset \mb R^{p_X}$ are random, the $d_{\rm stat}$ for a \matern kernel is $\m O\big(n^{1/(2+2\nu/p_X)}\big)$~\citep{tuo2020improved,chen2021fast},
whose rate is always less than $\sqrt{n}$.

For the leverage score approximation (which determines the quality score $\beta$), a descent algorithm ``BLESS'' \citep{rudi2018fast} can produce a constant order $\beta$ with only $\widetilde{\mathcal{O}}(nd)$ time complexity. 
Even better, if the kernel is additionally stationary, then the leverage scores can be estimated under a constant order relative approximation error with only $\widetilde{\mathcal{O}}(n)$ complexity, by utilizing the algorithm developed in \cite{chen2021fast}.

\section{Numerical Results}
\label{sec:exp}

In this section, we evaluate the empirical performance of our proposed \algoname{} method, with a focus on approximation error and computation time. 
We compare the \algoname{} method to Gaussian maps and vanilla importance sampling in various tasks, including (the sanity-checking task) approximating matrix product, spectral clustering for graph nodes (which involves sparse matrix objects), and kernel ridge regression (KRR; representative nonparametric regression models with intense matrix operations).
In general, we demonstrate that our method can reduce the projection dimension \emph{compared to} the vanilla importance sampling and thus benefit the total runtime. 
All experiments were implemented in Python with the NumPy / SciPy package and run with a single core of a server CPU (Intel Xeon-Gold 6248 @ 2.50GHZ) on Red Hat 4.8.

\begin{figure}[t]
\centering

\includegraphics[width=0.99\textwidth]{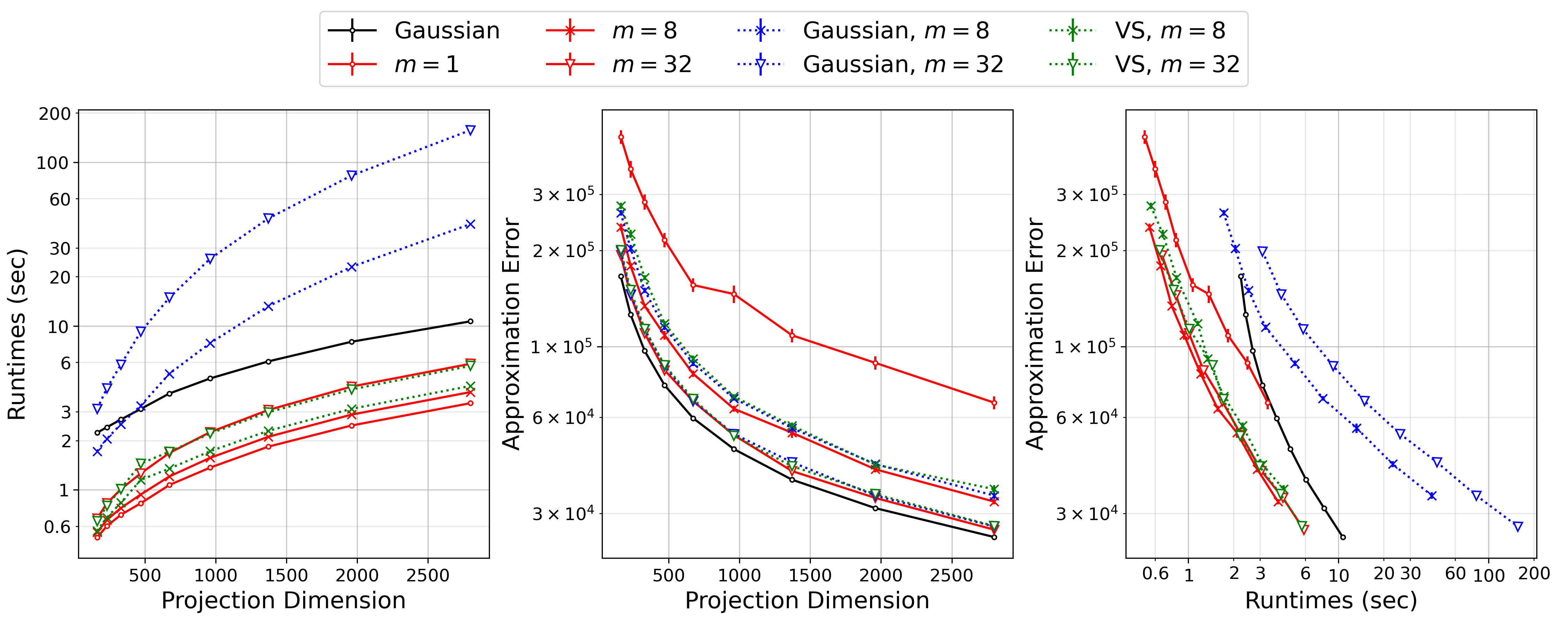}

\caption{
\textbf{Approximation error and runtime in approximating matrix product.}
Our method, \algoname{}, consistently achieves lower approximation error and runtime than the counterparts ``Gaussian'' and ``VS'' with the same parameter $m$. The superior performance is particularly evident for \algoname{} with $m=8$ (red curve with cross markers).
Error bars are included to quantify the variance and the performance gaps are significant.
}
\label{Fig:AMM}
\vspace{-3ex}
\end{figure}

\subsection{Sanity check: approximating large-scale matrix product}
\label{sec:exp_amm}

To provide thorough evaluation on the fundamental task, approximating matrix product, we extend the preliminary experiments in Section~\ref{sec:exp_amm_pre} and further examine the following methods: Gaussian sketching, regular sub-sampling, \algoname{} with $m=2,8,32$, its sub-Gaussian variant (introduced in Section~\ref{sec:sketching_composition}) with $m=2,8,32$, and very sparse random projection (VS) with $m=2,8,32$.
Following the settings in the VS literature~\cite{ahmad2022p}, we set the probability of an element in $\mtx \Pi$ being non-zero as $\frac{m}{n}$ for VS, so that the expected number of non-zero elements in VS is the same as the \algoname{} counterpart with the identical $m$.
(Notably, VS is a \emph{data-oblivious} method, and cannot improve with the sampling quality $\beta$.)

The matrices used for the matrix multiplication experiments were constructed as described in Section~\ref{sec:exp_amm_pre} (where $n=1000$), and we further tested the methods on matrices with larger sizes $n=2000, 4000$.
Due to space limit, here we only illustrate the performance for $n=4000$, while leave the complete results for all the sizes to Supplement~\ref{app:exp-amm};
the performance of each method is consistent among all the settings.

As shown in the last panel of Figure~\ref{Fig:AMM}, the ``error against runtime'' subplot, our proposed \algoname{} method continues to achieve the best trade-off between approximation error and runtime among all candidate methods. 
The sub-Gaussian variants of \algoname{} (labeled as ``Gaussian, $m=$'' along all the figures) and very sparse random projection (``VS, $m=$'') perform somewhere in between the other methods. 
They achieve comparable accuracy with \algoname{}, but their runtime efficiency is generally inferior to the proposed method due to either higher algorithmic complexity or slightly inefficient hardware implementations on top of sparse matrix multiplication. 

Figure~\ref{Fig:AMM} confirms that our proposed method can achieve lower approximation error than other methods for the same time budget in the fundamental task of approximating matrix product. 
We underscore that in the case where $m=8$, our method attains significantly lower approximation error (30 replicates make the gap statistically significantly) than its sub-Gaussian variants as well as VS (see the red, blue, and green curves with crosses in the last panel of Figure~\ref{Fig:AMM}; standard error bars are involved therein while short due to the large number of replicates).
In the following, we will demonstrate that our proposed \algoname{} method can improve the computational performance of statistical models as well.

\subsection{Spectral clustering}
\label{sec:exp_sc}

{
We further evaluate projection methods on the eigendecomposition-based statistical model, \emph{normalized spectral clustering} for graph nodes introduced by \citet[Section~2]{DBLP:conf/nips/NgJW01}. 
Different from the dense matrices in the previous matrix product experiments, we intentionally apply projection methods to \emph{sparse} graph Laplacian matrices in this experiment,
which brings new observations.

In performing spectral clustering, one needs to approximate the first $d$ eigenvectors (corresponding to the first $d$ eigenvalues in descending order) of the $n \times n$ signless normalized Laplacian matrix $\mtx{\tilde L} \defeq 2 \mtx{I} - \mtx{L}$. 
We specifically follow the steps in Section~\ref{sec:randomized-SVD}, 
compute the QR decomposition of the compressed Laplacian $\mtx{\Pi} \mtx{\tilde L} = \mtx{R}^\mathsf{T} \mtx Q^\mathsf{T}$ to obtain the $n \times d$ matrix $\mtx Q$, 
and then perform the eigendecomposition of the $d \times d$ matrix $\mtx Q^\mathsf{T} \mtx{\tilde L} \mtx Q$.
The main computation bottleneck lies in the decomposition operations \emph{after projection}, 
which underscores the necessity of controlling the projection dimension $d$.
}

\smallskip
\noindent \textbf{Experiment settings.}
We simulated graphs with sparse adjacency matrices using stochastic block models~\citep[SBM in short]{holland1983stochastic}.
To conduct a comprehensive evaluation, we varied the graph size $n$ as $5000, 10000$ and $15000$, and set the number of node groups $k$ to be $10$, $15$, and $20$ respectively.
(The results for $n=5000, 10000$ are deferred to Supplement~\ref{app:exp-sc} due to space limit.)

For SBM, we specified a connection matrix $\mtx P$ where $\mtx P_{ij}$ gives the probability of edges going from a node of group $i$ to another node of group $j$.
We set the diagonal elements of $\mtx P$ to be $0.3$ and the other elements to be $0.05$.
We also assumed a uniform multinomial distribution for group assignment, where each node has an equal probability of being assigned to any group.

We set a sequence of projection dimensions beforehand (from $250$ to $2250$) and in each round run randomized SVD (using a specific projection method) on a randomly generated graph with size $n$.
We then conduct spectral clustering based on the singular vectors 
extracted from randomized SVD of the signless normalized Laplacian matrices $\mtx{\tilde L}$.
To fairly compare the efficiency of different methods on \emph{sparse} matrices, 
we follow the same experiment settings as in Section~\ref{sec:exp_amm}:
we examine the same collection of methods and evaluate the runtime in obtaining the sparse matrix product $\mtx{\Pi} \mtx{\tilde L}$;
for the clustering accuracy metric, we adopt the normalized mutual information~\citep{JMLR:v11:vinh10a} between the clustering results and the ground truth as the metric.

\begin{figure}[t]
\centering
\includegraphics[width=0.85\textwidth]{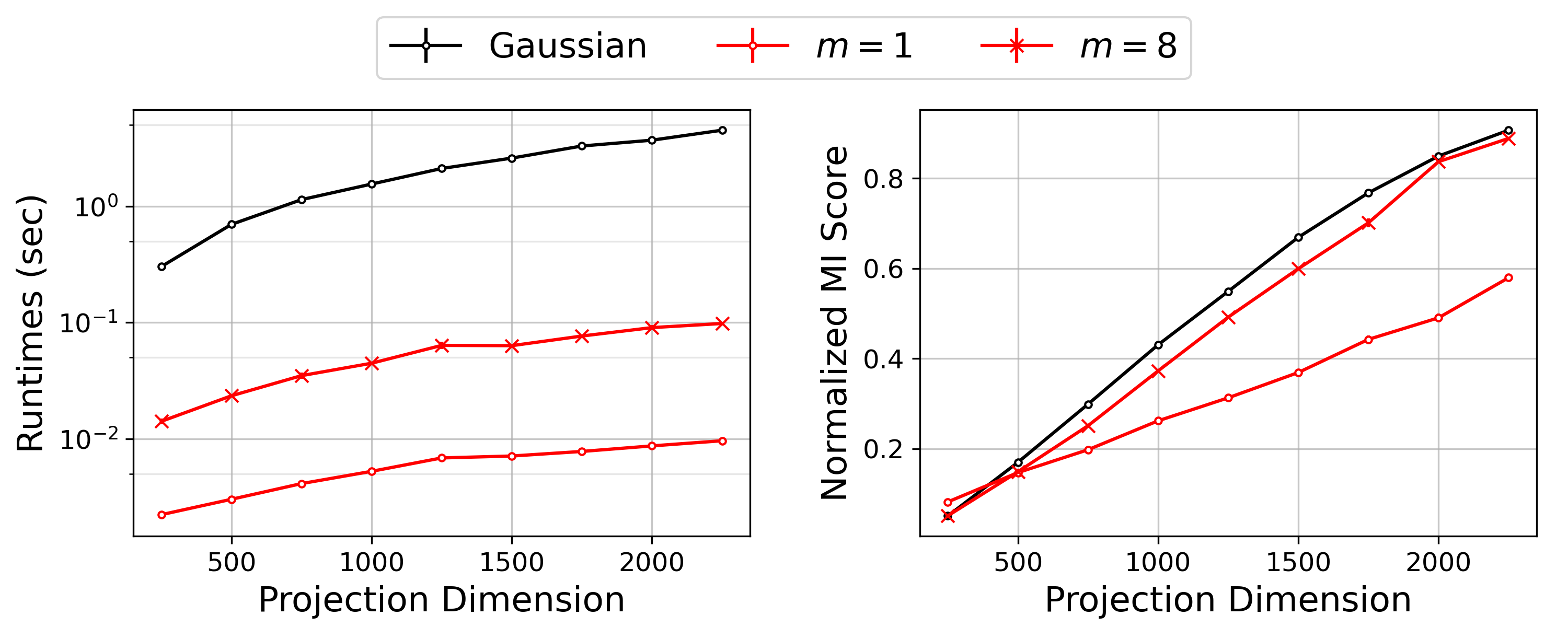}
\caption{\textbf{Runtime and clustering quality for representative projection methods in spectral clustering}. 
Regular sub-sampling with uniform importance (large $1/\beta$), marked with red circles, leads to the highest efficiency (the left panel) while the inferior clustering quality (the right panel) under the same projection dimension.
}
\label{Fig:sc0}
\vspace{-3ex}
\end{figure}

\smallskip
\noindent \textbf{Inapplicability of regular sub-sampling.} 
We first illustrate the inapplicability of regular sub-sampling in the scenario of spectral clustering, when the sampling quality is low (here we similarly employ uniform sampling).
At first sight, the projection procedure $\mtx{\Pi} \mtx{\tilde L}$ for sparse graph Laplacian matrices is highly efficient with regular sub-sampling, as shown in the left panel of Figure~\ref{Fig:sc0};
other methods need to further operate on sparse matrices in addition to drawing rows, 
which is naturally slower than the counterpart operations on dense matrices in Section~\ref{sec:exp_amm}.

However, the right panel of Figure~\ref{Fig:sc0} exhibits the suboptimal approximation performance of regular sub-sampling: 
it requires much higher projection dimension to attain the same performance as other projection methods,
while there will be a surge in the complexity of subsequent decomposition operations, which dominates the total computation cost.
For example, in Figure~\ref{Fig:sc0} the \algoname{} method with $m=8$ takes around 0.05 seconds to perform the $d=1000$-dim projection, while the following randomized SVD operations will cost around 1.6 sec.
To attain the same performance, Figure~\ref{Fig:sc0} suggests setting $d=1500$ for regular sub-sampling; 
although the projection step is highly efficient ($<0.01$ seconds),
the cost of randomized SVD will increase to around 4.0 sec for $d=1500$.
The total time cost for using regular sub-sampling with $d=1,500$ is much higher than \algoname{} with $d=1000$,
and therefore regular sub-sampling is inapplicable in this scenario.
We further investigate the practical time efficiency of other projection methods as follows.

\begin{figure}[t]
\centering
\includegraphics[width=0.99\textwidth]{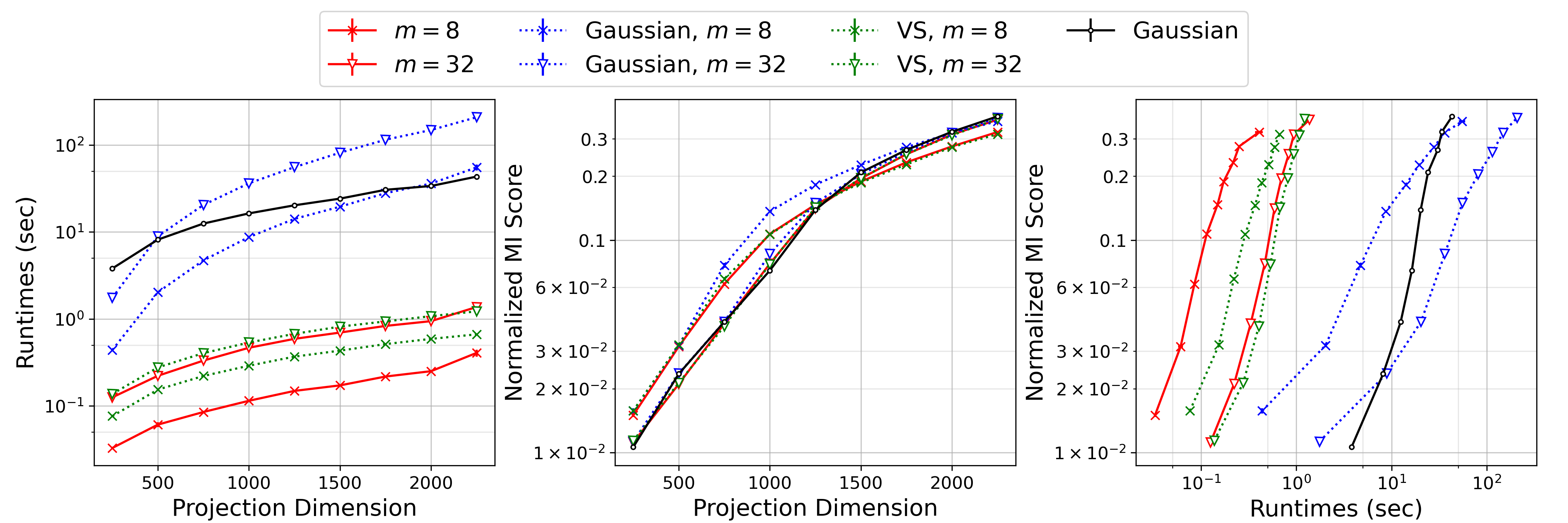}
\caption{\textbf{Runtime and clustering quality in spectral clustering}. 
Our method \algoname{}, when $m=8$ (the red curves with cross markers), consistently obtains higher normalized ML scores (the left panel) and less runtime (the middle panel) than the counterparts ``Gaussian'' and ``VS'' with $m=8$. A comprehensive comparison for the clustering quality and runtime is provided in the right panel.
}
\label{Fig:sc}
\vspace{-3ex}
\end{figure}

\smallskip
\noindent \textbf{Observations.} 
Figure~\ref{Fig:sc} verifies that \algoname{} with a medium value of $m=8$ can give comparable accuracy to subGaussian maps even with uniform sub-sampling distribution under the same projection dimension, and the projection cost is the lowest among other advanced methods. 
The significant gap in runtime between sparse methods (\algoname{} and VS) and the others (Gaussian and accumulative Gaussian sketching) indicates that our proposed methods are efficient and effective for sparse matrix objects (graph Laplacian matrix $\mtx{\tilde L}$ here) as well.

Due to the computational benefits and the capability of preserving graph information, overall our proposed \algoname{} attained improved trade-off between clustering quality and projection runtime. 
As shown in the last panel of Figure~\ref{Fig:sc}, the ``normalized MI score against runtime'' subplot, 
\algoname{} with $m=8$ (the red curve with cross markers) is the frontier among all the candidate methods.
The simulation here further verifies the prediction from our theoretical results (in Section~\ref{sec:randomized-SVD}) that a proper $m$ can lead to high-quality randomized SVD.

\subsection{Kernel ridge regression}
\label{Sec:exp_performance}

Following the evaluation frameworks above, we further test the accuracy and efficiency of different projection methods on KRR. 

\smallskip
\noindent \textbf{Experiment settings.}
We conducted the evaluation on three datasets downloaded from the UCI ML Repository \citep{Dua:2019},
including
\circled{1}~\texttt{RadiusQueriesAggregation} (denoted by RQA; \citealt{savva2018explaining, anagnostopoulos2018scalable}), 
\circled{2}~a dataset of physicochemical properties of protein tertiary structure (denoted by CASP; \citealt{CASP2013}), and 
\circled{3}~\texttt{PPGasEmission} (denoted by GAS; \citealt{KAYA2019}).
Here in the main text, we show the results on GAS, which contains $36,733$ data points and $10$ features, and defer the performance on the other two datasets to Supplement~\ref{app:exp-krr}.

We mainly follow the experiment settings in \cite{chen2021accumulations}.
In each run, we randomly select $15,000$ samples for training 
and estimate the testing errors on another random subset ($20\%$ of the original dataset) $(\mtx X_{\text{new}}, \mtx Y_{\text{new}})$ with size $n_{\text{new}}$, which is unseen in training. 
We start by normalizing the features to have variance $1$ in the randomly selected dataset, and then compute the empirical kernel matrix using the \matern kernel. 
After cross-validation for the smallest testing error by the original KRR, the smoothness parameter $\nu$ in the \matern kernel is set as $1$. 
Denoting the number of features as $p_X$, we set the regularization parameter $\lambda$ of KRR as $0.9 \cdot n^{-\paren{3+p_X}/\paren{3+2p_X}}$, and set the projection dimension $d$ ranging from $250$ to $2,250$. 

The candidate methods include Gaussian sketching, very sparse random projection, the classical \nystrom method (uniform sampling with $m=1$), and our \algoname{} method, all implemented with uniform sampling (if applicable) for fair comparison. We set the accumulative parameter $m=4$ for \algoname{} / its subGaussian variant / VS, which allows all the methods to achieve comparable accuracy (except for the classical \nystrom method) to subGaussian maps.

\begin{figure}[t]
\centering
\includegraphics[width=0.99\textwidth]{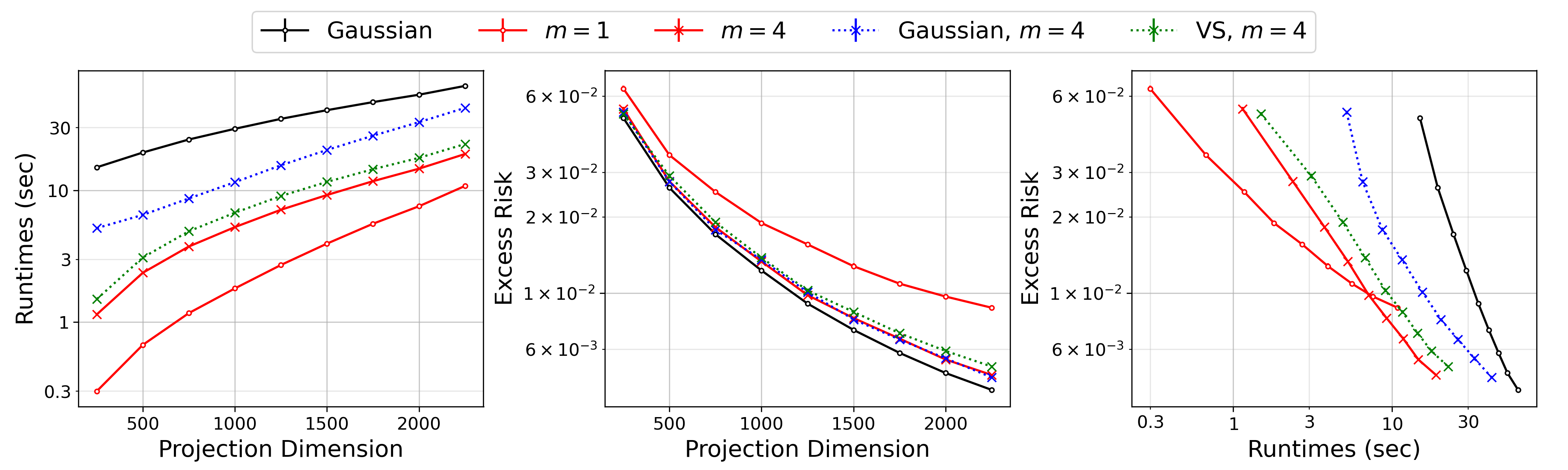}
\caption{\textbf{Runtime and excess risk in KRR}. 
The first two panels illustrate how runtime and excess risk change with projection dimensions. 
Our method, \algoname{}, with $m=4$ (red curve with cross markers), consistently achieves both low runtime and high efficiency;
in the last panel, \algoname{} gradually surpasses regular sampling ($m=1$, the red curve with circle markers) when the required excess risk in KRR is below $1 \times 10^{-2}$.}
\label{Fig:krr_datasets}
\vspace{-3ex}
\end{figure}

Overall, the sketched KRR estimator $\hat{f_S}$ \cref{eqn:krr-fS} approximates the original KRR estimator $\hat{f_n}$ \eqref{eqn:krr-fn}, and a natural choice of the metric in the KRR task is the prediction error $\frac{1}{n_{\text{new}}} \|\hat{f_n}(\mtx X_{\text{new}}) - \mtx Y_{\text{new}}\|^2$.
However, the scale of the approximation error (of our numerical interest) $\frac{1}{n_{\text{new}}} \|\hat{f_S}(\mtx X_{\text{new}}) - \hat{f_n}(\mtx X_{\text{new}})\|^2$ is usually smaller than the prediction error, 
and we choose the excess risk (which is dominated by the approximation error)
\begin{align*}
\frac{1}{n_{\text{new}}} \norm{\hat{f_S}(\mtx X_{\text{new}}) - \mtx Y_{\text{new}}}^2 - \frac{1}{n_{\text{new}}} \norm{\hat{f_n}(\mtx X_{\text{new}}) - \mtx Y_{\text{new}}}^2
\end{align*}
as the KRR accuracy metric for better visualization of the trend. 
All the results reported in Figure~\ref{Fig:krr_datasets} are averaged over 20 replicates.

\smallskip
\noindent \textbf{Observations.} 
Those experiments demonstrate that in practice, \algoname{} with medium values of $m$ can substantially improve the accuracy of the classical \nystrom method ($m=1$) in KRR, even with uniform sub-sampling and large $1/\beta$.
The computational cost of \algoname{} is low compared to other advanced methods. 
Figure~\ref{Fig:krr_datasets}, along with Figure~\ref{Fig:app-krr} in the supplement, illustrates that on all the three datasets, \algoname{} (the red curve with crosses) can attain comparable accuracy with subGaussian maps and slightly exceed VS. 
While classical \nystrom methods ($m=1$) are the most efficient under the same projection dimension, they fail to obtain the same accuracy as other methods, especially when the projection dimension grows---its asymptotic excess risk versus runtime trade-off will ultimately be surpassed by \algoname{}. 
Overall, on those datasets \algoname{} provides a favorable trade-off between statistical accuracy and computational efficiency among all candidate projection methods.

\section{Conclusion}
We introduce \algoname{}, a sampling-based random projection method that 
aggregates multiple re-scaled randomly signed sub-sampling matrices to reduce the projection dimension.
This design facilitates follow-up extensive matrix operations in statistical computation. 
We prove that the new method can provide a spectral norm guarantee for approximating matrix product with a projection dimension as low (up to poly-log factors) as potent sub-Gaussian maps. 
Additionally, we establish the effectiveness of \algoname{} in approximating a PSD empirical kernel matrix. 
We provide a theoretical analysis of its application to statistical computation, including eigendecomposition (via randomized SVD) and \nystrom approximation for fast kernel ridge regression. 
For empirical evaluation, we conduct comprehensive experiments to demonstrate the efficacy of our method in aforementioned learning tasks.

\clearpage
\bibliographystyle{mldraft}
\bibliography{accu}

\clearpage
\appendix

\section{More on empirical experiments}
\label{app:exp}

We provide additional experiment results in this section, which are mainly summarized in the corresponding figures.

\subsection{Supplementary information for the experiments on approximating matrix multiplication}
\label{app:exp-amm}

\begin{figure}[ht]
\centering
\subfigure[n = 1,000]{
\includegraphics[width=0.85\textwidth]{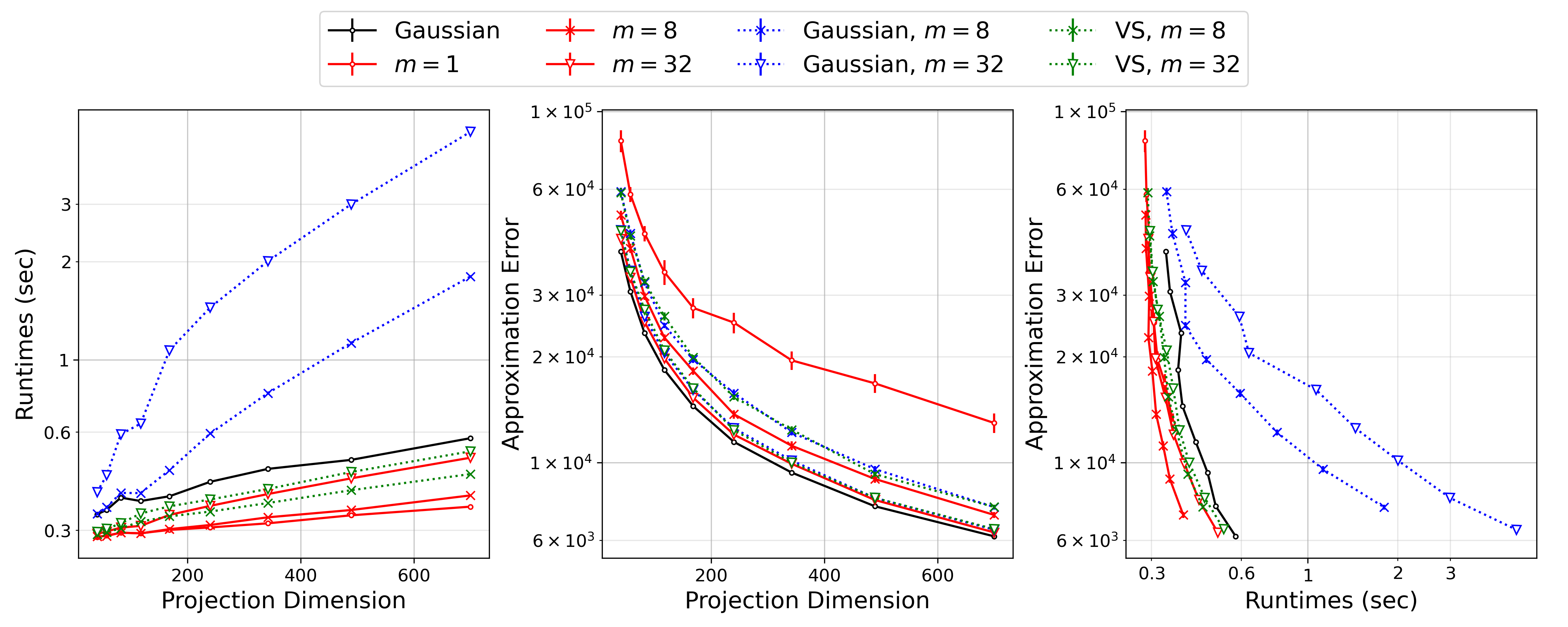}
}
\quad
\subfigure[n = 2,000]{
\includegraphics[width=0.85\textwidth]{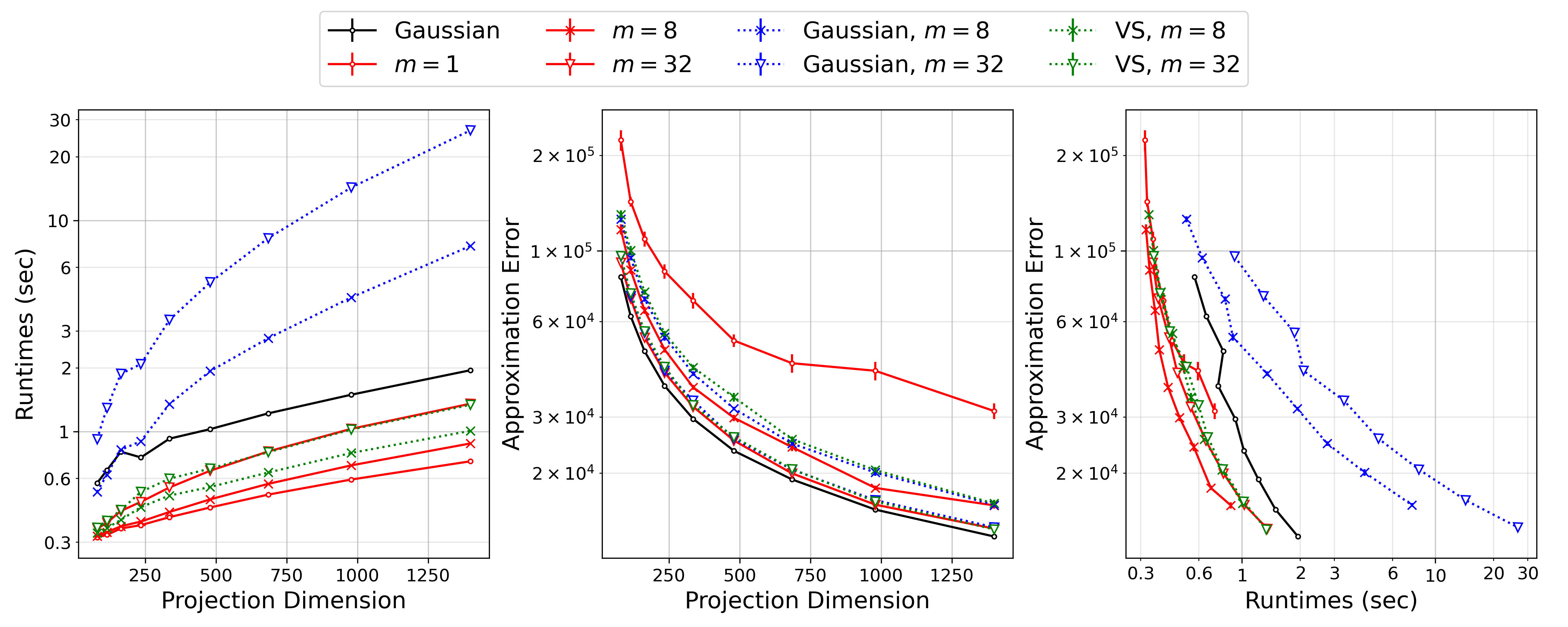}
}
\quad
\subfigure[n = 4,000]{
\includegraphics[width=0.85\textwidth]{exp1_2_accu_tradeoff_bar2.png}
}
\caption{
Accuracy v.s.\ efficiency trade-off in approximating large-scale matrix product. 
}
\label{fig:app-AMM}
\end{figure}

We extend the preliminary experiments in Section~\ref{sec:exp_amm_pre} and further examine the following methods: Gaussian sketching, sub-sampling sketching, \algoname{} with $m=2,8,32$, its sub-Gaussian variant (introduced in Section~\ref{sec:sketching_composition}) with $m=2,8,32$, and very sparse random projection (VS) with $m=2,8,32$.
Following the settings in the VS literature~\cite{ahmad2022p}, we set the probability of an element in $\mtx \Pi$ being non-zero as $\frac{m}{n}$ for VS, so that the expected number of non-zero elements in VS is the same as the \algoname{} counterpart with the identical $m$.
We set the matrix sizes as $n=1000, 2000, 4000$ and showed the experiment results in Figure~\ref{fig:app-AMM}.

\subsection{Supplementary experiments on spectral clustering}
\label{app:exp-sc}

\begin{figure}[ht]
\centering
\subfigure[n = 5,000]{
\includegraphics[width=0.85\textwidth]{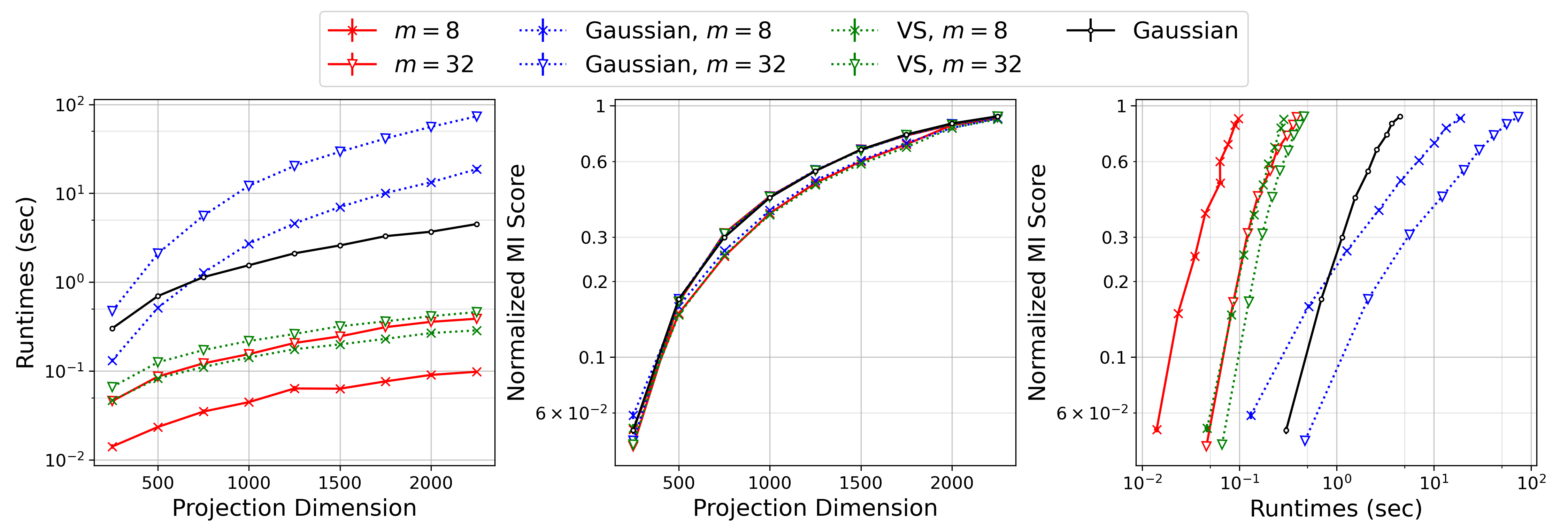}
}
\quad
\subfigure[n = 10,000]{
\includegraphics[width=0.85\textwidth]{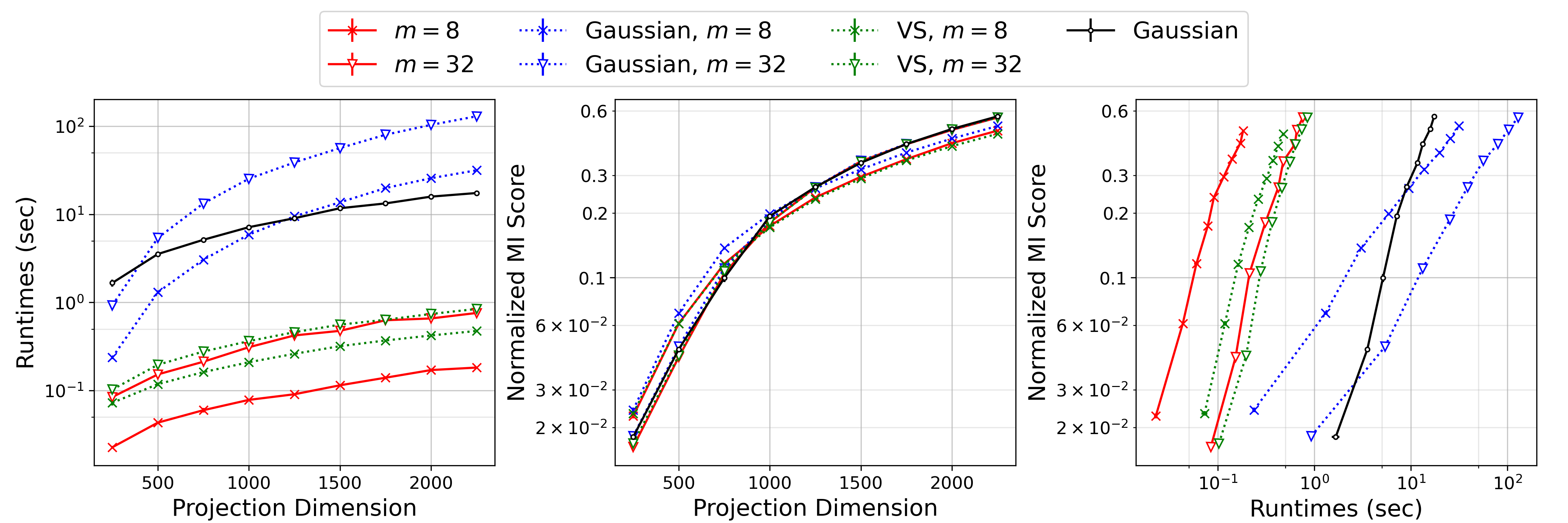}
}
\quad
\subfigure[n = 15,000]{
\includegraphics[width=0.85\textwidth]{exp2_tradeoff_bar_revise2.png}
}
\caption{Trade-off between accuracy and efficiency in spectral clustering. 
}
\label{Fig:app-sc}
\end{figure}

We manually generated graphs with sparse adjacency matrices using stochastic block models~\citep[SBM in short]{holland1983stochastic}.
To conduct a comprehensive evaluation, we varied the graph size $n$ as $5000, 10000$ and $15000$, and set the number of node groups $k$ to be $10$, $15$, and $20$ respectively.
The experiment results are provided in Figure~\ref{Fig:app-sc}.

\subsection{Supplementary experiments on kernel ridge regression}
\label{app:exp-krr}

\begin{figure}[t]
\centering
\subfigure[RQA]{
\includegraphics[width=0.85\textwidth]{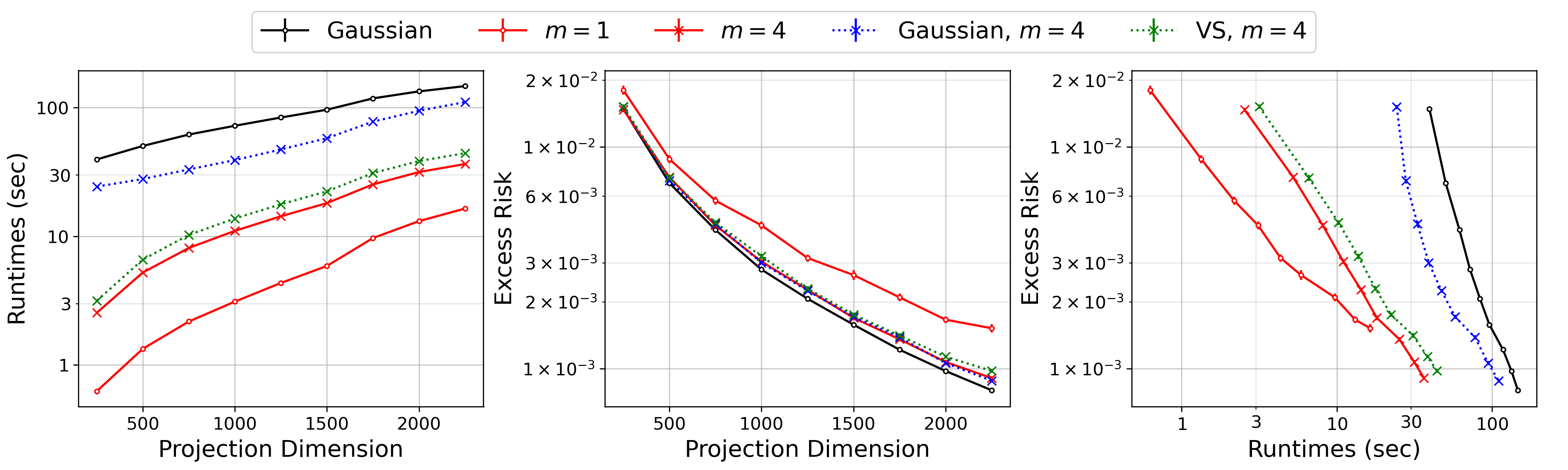}
}
\quad
\subfigure[CASP]{
\includegraphics[width=0.85\textwidth]{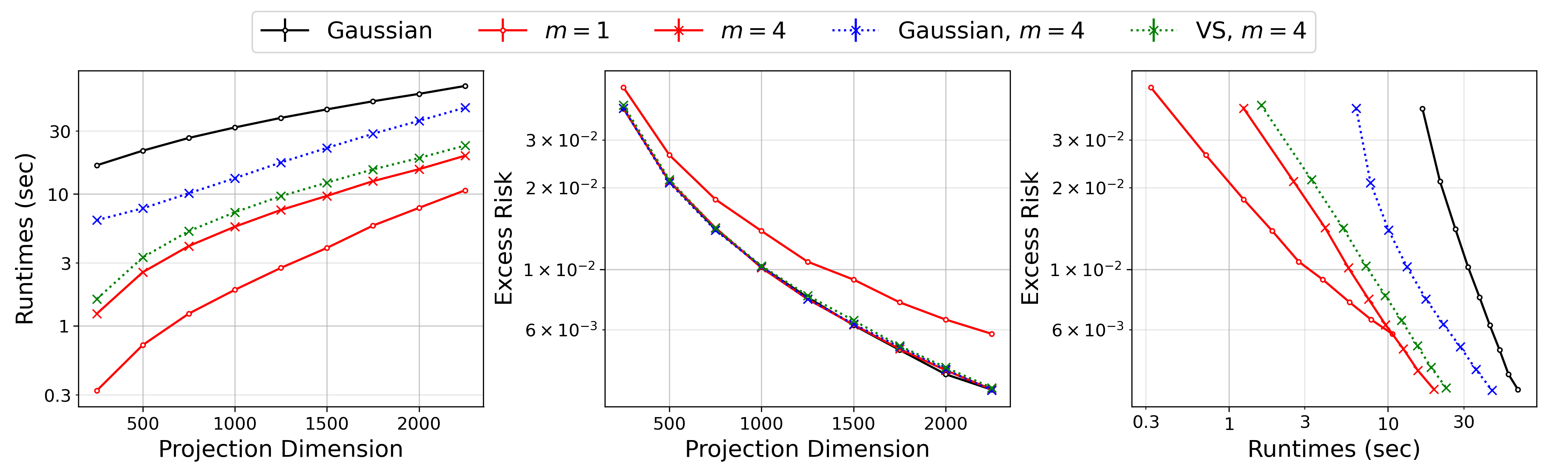}
}
\quad
\subfigure[GAS]{
\includegraphics[width=0.85\textwidth]{exp3_revise_2.png}
}
\caption{Trade-off between accuracy and efficiency in KRR.}
\label{Fig:app-krr}
\end{figure}

We conduct the evaluation on three datasets downloaded from the UCI ML Repository \citep{Dua:2019}: 
\texttt{RadiusQueriesAggregation} (denoted by RQA; \citealt{savva2018explaining, anagnostopoulos2018scalable}), a data set of physicochemical properties of protein tertiary structure (denoted by CASP; \citealt{CASP2013}), and \texttt{PPGasEmission} (denoted by GAS; \citealt{KAYA2019}).
For those datasets, RQA contains $200,000$ data points and $4$ features;
CASP contains $45,730$ data points and $9$ features;
GAS contains $36,733$ data points and $10$ features.
The evaluation results are summarized in Figure~\ref{Fig:app-krr}.

\section{Useful facts: matrix concentration inequalities}

The following theorems are mainly taken from a tutorial \citep[Theorems~1.1, 1.4, and~1.6]{tropp2012user} for the reader's convenience. 
For the last Vector Hoeffding inequality, we adapt the vector Bernstein inequality~\citep[Lemma~18]{kohler2017sub} and provide the proof in this section for self-containedness.

\begin{theorem}[Matrix Rademacher Series: Rectangular Case]
\label{thm:intro-rad-rect}
Consider a finite sequence $\{ \mtx{X}_k \}$ of independent, random, self-adjoint matrices with dimension $n$.
Assume that each random matrix satisfies
$$
\mtx{X}_k \psdge \mtx{0}
\quad\text{and}\quad
\lambda_{\max}( \mtx{X}_k ) \leq R,
\quad \forall k~\text{almost surely}.
$$
Define
$$
\mu_{\min} \defeq \lambda_{\min} \left( \sum\nolimits_k \Expect\brkt {\mtx{X}_k} \right) \quad \text{and} \quad 
\mu_{\max} \defeq \lambda_{\max}\left( \sum\nolimits_k \Expect\brkt {\mtx{X}_k} \right).
$$
Then
\begin{align*}
\Prob{ \lambda_{\min}\left( \sum\nolimits_k \mtx{X}_k \right) \leq (1 - \eta) \mu_{\min} }
	&\leq n \cdot \left[ \frac{\econst^{-\eta}}{(1 - \eta)^{1 - \eta}} \right]^{\mu_{\min}/R}, \quad \text{for~} \eta \in [0, 1], \text{and} \\
\Prob{ \lambda_{\max}\left( \sum\nolimits_k {X}_k \right) \geq (1 + \eta) \mu_{\max} }
	&\leq n \cdot \left[ \frac{\econst^{\eta}}{(1+\eta)^{1 + \eta}} \right]^{\mu_{\max}/R}, \quad \text{for~} \eta \geq 0.
\end{align*}
\end{theorem}
\begin{theorem}[Matrix Chernoff]
\label{thm:intro-chernoff}
Consider a finite sequence $\{ \mtx{X}_k \}$ of independent, random, self-adjoint matrices with dimension $n$.
Assume that each random matrix satisfies
$$
\mtx{X}_k \psdge \mtx{0}
\quad\text{and}\quad
\lambda_{\max}( \mtx{X}_k ) \leq R,
\quad \forall k~\text{almost surely}.
$$
Define
$$
\mu_{\min} \defeq \lambda_{\min} \left( \sum\nolimits_k \Expect\brkt {\mtx{X}_k} \right) \quad \text{and} \quad 
\mu_{\max} \defeq \lambda_{\max}\left( \sum\nolimits_k \Expect\brkt {\mtx{X}_k} \right).
$$
Then
\begin{align*}
\Prob{ \lambda_{\min}\left( \sum\nolimits_k \mtx{X}_k \right) \leq (1 - \eta) \mu_{\min} }
	&\leq n \cdot \left[ \frac{\econst^{-\eta}}{(1 - \eta)^{1 - \eta}} \right]^{\mu_{\min}/R}, \quad \text{for~} \eta \in [0, 1], \text{and} \\
\Prob{ \lambda_{\max}\left( \sum\nolimits_k {X}_k \right) \geq (1 + \eta) \mu_{\max} }
	&\leq n \cdot \left[ \frac{\econst^{\eta}}{(1+\eta)^{1 + \eta}} \right]^{\mu_{\max}/R}, \quad \text{for~} \eta \geq 0.
\end{align*}
\end{theorem}
\begin{theorem}[Matrix Bernstein: Square case] \label{thm:intro-bernstein}
Consider a finite sequence $\{ \mtx{X}_i \}$ of independent, random, self-adjoint matrices with dimension $n$.  Assume that each random matrix satisfies
$$
\Expect\brkt {\mtx{X}_i} = \mtx{0}
\quad\text{and}\quad
\matnorm{\mtx{X}_i} \leq R
\quad\text{almost surely}.
$$
Then, for all $t \geq 0$,
$$
\Prob{ \matnorm{\sum\nolimits_i \mtx{X}_i} \geq t }
	\leq 2n \cdot \exp\left( \frac{-t^2/2}{\sigma^2 + Rt/3} \right)
	\quad\text{where}\quad
	\sigma^2 \geq \matnorm{ \sum\nolimits_i \Expect \brkt{\mtx{X}_i^2} }.
$$
\end{theorem}
\begin{theorem}[Matrix Bernstein with Bounded Norms: Rectangular Case] \label{thm:intro-bernstein-rect}
Consider a finite sequence $\{ \mtx{Y}_i \}$ of independent, random matrices with dimensions $n_1 \times n_2$.  Assume that each random matrix satisfies
$$
\Expect\brkt {\mtx{Y}_i} = \mtx{0}
\quad\text{and}\quad
\matnorm{ \mtx{Y}_i } \leq R
\quad\text{almost surely}.
$$
Define
$$
\sigma^2 \defeq \max\left\{ 
	\matnorm{ \sum\nolimits_i \Expect\brkt{\mtx{Y}_i \mtx{Y}_i^\adj} }, \
	\matnorm{ \sum\nolimits_i \Expect\brkt{\mtx{Y}_i^\adj \mtx{Y}_i} }
\right\}.
$$
Then, for all $t \geq 0$,
$$
\Prob{ \matnorm{ \sum\nolimits_i \mtx{Y}_i } \geq t }
	\leq (n_1 + n_2) \cdot \exp\left( \frac{-t^2/2}{\sigma^2 + Rt/3} \right).
$$
\end{theorem}
\begin{theorem}[Vector Bernstein] \label{thm:intro-bernstein-vect}
Consider a finite sequence $\{ \mtx{Y}_i \}$ of independent, random vectors with dimensions $p$.  
Assume that each random vector satisfies
$$
\Expect\brkt {\mtx{Y}_i} = \mtx{0}
\quad\text{and}\quad
\norm{ \mtx{Y}_i } \leq R
\quad\text{almost surely}.
$$
Define
$$
\sigma^2 \defeq \sum\nolimits_i \Expect\brkt{\mtx{Y}_i^\mathsf{T} \mtx{Y}_i}.
$$
Then, for all $t \geq 0$,
$$
\Prob{ \norm{ \sum\nolimits_i \mtx{Y}_i } \geq \sigma + t}
	\leq \exp\paren{-\frac{t^2}{8 \sigma^2 + \frac{4R}{3} t }}.
$$
\end{theorem}

\begin{proof}
We will construct a Doob martingale to attain the claim.
We define $\mtx S_n = \sum\nolimits_i \mtx{Y}_i$, $D_1 = \Expect\brkt{ \|\mtx S_n\| \mid \mtx Y_1} - \Expect\brkt{ \|\mtx S_n\| }$, and 
$$
D_i \defeq \Expect\brkt{ \|\mtx S_n\| \mid \mtx Y_1, \dots, \mtx Y_i } - \Expect\brkt{ \|\mtx S_n\| \mid \mtx Y_1, \dots, \mtx Y_{i-1}}, 
\quad \forall i \geq 2.
$$
We note $D_i$'s constitute a martingale difference sequence, and $\sum\nolimits_i D_i = \|\mtx S_n\| - \Expect\brkt{ \|\mtx S_n\| }$ is the quantity of interest.

We first show $D_i$'s are bounded. Specifically, 
\begin{align*}
D_i &= \Expect\brkt{ \|\mtx S_n - \mtx Y_i + \mtx Y_i \| \mid \mtx Y_1, \dots, \mtx Y_i } - \Expect\brkt{ \|\mtx S_n - \mtx Y_i + \mtx Y_i \| \mid \mtx Y_1, \dots, \mtx Y_{i-1}} \\
&\leq \Expect\brkt{ \|\mtx S_n - \mtx Y_i\| + \|\mtx Y_i\| \mid \mtx Y_1, \dots, \mtx Y_i } - \Expect\brkt{ \|\mtx S_n - \mtx Y_i\| - \| \mtx Y_i \| \mid \mtx Y_1, \dots, \mtx Y_{i-1}} \\
&= \|\mtx Y_i\| + \Expect\brkt{ \| \mtx Y_i \| } \leq 2 R,
\end{align*}
and we similarly obtain $D_i \geq -2R$.
Utilizing $|D_i| \leq \|\mtx Y_i\| + \Expect\brkt{ \| \mtx Y_i \| }$,
we can further bound its (conditional) second moment:
$\Expect [D_i^2 \mid \mtx Y_1, \dots, \mtx Y_{i-1}] \leq \Expect \|\mtx Y_i\|^2 + 3 \Expect^2 \|\mtx Y_i\| \leq 4 \Expect \|\mtx Y_i\|^2$.

We can specify the decaying rate of $\Prob{\sum\nolimits_i D_i > t}$ through bounding
$$
\Expect\brkt{\exp\paren{ \lambda \sum\nolimits_i D_i }} = 
\Expect\brkt{\exp\paren{ \lambda \sum_{i=1}^{n-1} D_i } \Expect\brkt{ \exp( \lambda D_n ) \mid \mtx Y_1, \dots, \mtx Y_{n-1} } }.
$$
Here, we note in the standard proof of the Bernstein inequality,
$$
\Expect\brkt{ \exp( \lambda D_n ) \mid \mtx Y_1, \dots, \mtx Y_{n-1} } \leq \exp\paren{\frac{\lambda^2 \Expect [D_n^2 \mid \mtx Y_1, \dots, \mtx Y_{n-1}]}{2 (1 - \frac{\lambda}{3} 2 R ) }},
$$ 
and we have
$$
\Expect\brkt{\exp\paren{ \lambda \sum\nolimits_i D_i }} \leq
\Expect\brkt{\exp\paren{ \lambda \sum_{i=1}^{n-1} D_i } } \cdot \exp\paren{\frac{\lambda^2 4 \Expect \|\mtx Y_i\|^2}{2 (1 - \frac{\lambda}{3} 2 R ) }}.
$$
By similarly peeling off other terms, we have
$$
\Expect\brkt{\exp\paren{ \lambda \sum\nolimits_i D_i }} \leq
\exp\paren{\frac{\lambda^2 4 \sum\nolimits_i \Expect \|\mtx Y_i\|^2}{2 (1 - \frac{\lambda}{3} 2 R ) }}
= \exp\paren{\frac{\lambda^2 4 \sigma^2}{2 (1 - \frac{\lambda}{3} 2 R ) }}.
$$
Again, through the standard proof of the Bernstein inequality, we have
\begin{align*}
\Prob{\sum\nolimits_i D_i > t} &\leq \exp\paren{-\frac{t^2}{2 (4 \sigma^2 + \frac{2R}{3} t) }} \\
\quad \Leftrightarrow \quad
\Prob{\|\mtx S_n\| > \Expect\brkt{ \|\mtx S_n\| } + t} &\leq \exp\paren{-\frac{t^2}{8 \sigma^2 + \frac{4R}{3} t }}.
\end{align*}
The last step is to relax the mean $\Expect\brkt{ \|\mtx S_n\| }$ as $\Expect^{\frac12}\brkt{ \|\mtx S_n\|^2 }$.
We note
$$
\Expect \brkt{ \|\mtx S_n\|^2 } = \Expect \brkt{ \sum\nolimits_i \|\mtx Y_i\|^2} = \sigma^2,
$$
considering the interaction terms are mean-zero.
Finally, we have
$$
\Prob{\|\mtx S_n\| > \sigma + t} \leq \Prob{\|\mtx S_n\| > \Expect\brkt{ \|\mtx S_n\| } + t} \leq \exp\paren{-\frac{t^2}{8 \sigma^2 + \frac{4R}{3} t }}.
$$
The claim is then proved.
\end{proof}

\section{Revisiting selected random projection methods}

The focus of this paper is the handling of suboptimal sampling probabilities, while it is also formally related to randomized sketching methods due to the shared formulation as a random sketching matrix $\mtx{\Pi}$.
We revisit the development of generic randomized sketching methods in this section for a clearer context, which especially streamlines the statement of the implicit connection between two representative sketching methods in Section~\ref{sec:def}.

\subsection{Literature review of data-oblivious projection methods}
\label{sec:oblivious-methods}

Various methods have been developed to construct a data-oblivious random sketching matrix $\mtx{\Pi}$ for \emph{different purposes}.
\ding{182}~One of the most common applications of randomized sketching is to implement the \emph{Johnson--Lindenstrauss} (JL) \emph{transform}~\citep{johnson1984extensions}, 
a powerful tool in dimension reduction.
The JL transform is featured with a random sketching matrix $\mtx{\Pi} \in \mb R^{d\times n}$ so that for any fixed $\mtx z \in \mathbb{R}^n$ with $\|\mtx z\|=1$, $\|\mtx \Pi \mtx z\|$ is close to $1$ with high probability.
The most straightforward construction of the sketching matrix sets the entries in $\mtx \Pi$ as independent and identically distributed (i.i.d.) sub-Gaussian random variables, with representative examples including sub-Gaussian maps \citep{vershynin2010introduction, DBLP:journals/siamrev/HalkoMT11} and very sparse random projection \citep[entries are zero with a certain probability;][]{achlioptas2001database, li2006very, ahmad2022p}.
To accelerate the computation of $\mtx{\Pi z}$, one can also set $\mtx \Pi$ as a Fast JL transform~\citep{ailon2009fast}, such as the sub-sampled randomized Hadamard transform (SRHT)~\citep{sarlos2006improved, ailon2006approximate, lu2013faster, yang2017randomized},
or a sparse JL transform~\citep{dasgupta2010sparse, kane2014sparser} such as Count Sketch~\citep{charikar2004finding}.

A more ambitious application, \textit{oblivious subspace embedding} (OSE) further extends the aforementioned JL transform.
Specifically, OSE requires that for any orthonormal matrix $\mtx U \in \mb R^{n \times r}$ ($\mtx U^\mathsf{T} \mtx U = \mtx I_r$, and thus $\mtx U$ is the basis for a subspace), 
$\matnorm{(\mtx{\Pi U})^\mathsf{T} (\mtx{\Pi U}) - \mtx I_r}$\footnote{Throughout this paper, $\matnorm{\mtx A}$ represents the matrix operator norm with respect to $\|\cdot\|$.} should be upper bounded with high probability.
To achieve the desired OSE property while quickly computing the products $\mtx{\Pi A}, \mtx{\Pi B}$, variants of the Fast JL transform (such as SRHT) and sparse OSE~\citep{nelson2013osnap, meng2013low, cohen2016nearly, clarkson2017low} can be utilized.
{It is worth noting that JL transforms and OSE impose strong requirements on the sketching methods, as they relate to the worst-case error for arbitrary vectors/subspaces. 
In this context, all the methods mentioned above are \emph{data-oblivious} methods, where the probability distribution over sketching matrices is independent of input data.
}

\subsection{Comparison of projection matrix sparsity for data-adaptive / data-oblivious approaches}
\label{sec:compare-sparsity}

In contrast to most data-oblivious methods discussed earlier, we focus our attention on data-adaptive sampling-based methods (and do not pursue the JL and OSE properties). 
From a computing system perspective, sub-sampling sketching is fast to compute, as the underlying indexing operations are highly efficient, even on GPUs. 
Moreover, if we have access to the optimal sub-sampling probabilities, the number of non-zero elements in the sketching matrix can be effectively reduced to magnitudes comparable to the stable ranks of the target matrices~\citep{drineas2006fast}. 
For instance, the stable rank of a Gaussian kernel matrix is poly-logarithmic with respect to the sample size (also matrix size) $n$~\citep{chen2021fast}.
The computational bottleneck, however, lies in obtaining the optimal sub-sampling probabilities, which is often as time-consuming as the exact matrix multiplication~\citep{alaoui2015fast}. If only a rough estimate of the optimal sub-sampling probabilities is available due to a limited runtime budget, the projection dimension needed by the sub-sampling-based method can be much larger than that of the above data-oblivious methods (notice the factor $1/\beta$ in Theorem~\ref{thm:amm_asym} when setting $m=1$); this, in turn, increases the downstream computation cost. To address this issue, we endeavor in this work to improve sub-sampling sketching under the case of suboptimal sampling probabilities.

The matrix sparsity of sampling-based projection methods can be made smaller than the lower bound $\Omega(n \log n)$~\citep{kane2014sparser} previously proved to be required for the Johnson--Lindenstrauss (JL) transform (\citealt{johnson1984extensions}; introduced in Section~\ref{sec:related}), 
which require that for \emph{any} fixed vector $\mtx z \in \mathbb{R}^n$ with $\|\mtx z\|=1$\footnote{Throughout this paper, $\|\mtx z\|$ represents the vector $\ell_2$ norm.}, 
a random projection matrix $\mtx{\Pi} \in \mb R^{d\times n}$ enforces $\|\mtx \Pi \mtx z\|$ close to $1$ with high probability;
instead, sampling-based methods target at the approximation confined to specific matrices with low stable ranks.

\section{Proof of Theorem~\ref{thm:amm_asym}}
\label{sec:amm_asym}

\begin{proof}
The core idea in the proof is to utilize the rectangular matrix Bernstein inequality (Theorem~\ref{thm:intro-bernstein-rect}    ) 
about the concentration of sums of independent matrices around their expectations.
We let $\mtx{X}_i \defeq \mtx{A}^\mathsf{T} \mtx{\Pi}_{(i)} (\mtx{\Pi}_{(i)})^\mathsf{T} \mtx{B} - \frac1d \mtx{A}^\mathsf{T} \mtx{B}, \forall i=1,\dots,d$, and note they are independent, zero-mean matrices as required.
We first give the upper bound $R$ for the spectral norm of $\mtx{X}_i$'s, and it will be $R_A R_B + \frac1d \matnorm{\mtx{A}}\matnorm{\mtx{B}}$,
where $R_A \defeq \max_i \|\mtx{A}^\mathsf{T} \mtx{\Pi}_{(i)}\|, R_B \defeq \max_i \|\mtx{B}^\mathsf{T} \mtx{\Pi}_{(i)}\|$.
Technically, $R_A, R_B$ are proportional to the accumulation parameter $m$, which is too large;
we apply a trick here to utilize the high probability bounds of $R_A, R_B$ with the vector Hoeffding inequality (Theorem~\ref{thm:intro-bernstein-vect}) and the union bound.

The construction of the sketching matrix $\mtx{\Pi}$ implies that 
\begin{align*}
\mtx{\Pi}_{(i)} = \sum_{k=1}^m \frac{1}{\sqrt{m}} \mtx{\Pi}_{k, (i)}, \quad \forall i \in [d],
\end{align*}
where $\mtx{\Pi}_k$ represents a $n$-by-$d$ randomly signed sub-sampling sketching matrix with the scheme $\{p_j\}_{j=1}^n$.
For $\frac{1}{\sqrt{m}} \mtx{A}^\mathsf{T} \mtx{\Pi}_{k, (i)}$, we can follow the proof by \cite{alaoui2015fast} and bound its operator norm $R_A^0$ as
\begin{align*}
\|\frac{1}{\sqrt{m}} \mtx{A}^\mathsf{T} \mtx{\Pi}_{k, (i)}\| 
&\leq \max_{j} \frac{1}{\sqrt{m d}} \frac{\|\mtx{A}_{(j)}\|}{\sqrt{p_j}} \\
&\leq \matnorm{\mtx{A}} \sqrt{\frac{\sum_{j=1}^{n} \max\{\|\mtx{A}_{(j)}\|^2 / \matnorm{\mtx{A}}^2, \|\mtx{B}_{(j)}\|^2 / \matnorm{\mtx{B}}^2\}}{m d \beta}} \\
&\leq \matnorm{\mtx{A}} \sqrt{\frac{\matnorm{\mtx{A}}_{\rm F}^2 / \matnorm{\mtx{A}}^2 + \matnorm{\mtx{B}}_{\rm F}^2 / \matnorm{\mtx{B}}^2\}}{m d \beta}} 
\leq \sqrt{\frac{2s\matnorm{\mtx{A}}^2}{m d \beta}}.
\end{align*}
Its ``variance'' 
\begin{align*}
(\sigma_A^0)^2 \defeq \max\left\{ 
	\matnorm{ \frac1m \sum_{k=1}^m \Expect\brkt{ \mtx{A}^\mathsf{T} \mtx{\Pi}_{k, (i)} \mtx{\Pi}_{k, (i)}^\mathsf{T} \mtx{A}} }, \
	\matnorm{ \frac1m \sum_{k=1}^m \Expect\brkt{\mtx{\Pi}_{k, (i)}^\mathsf{T} \mtx{A} \mtx{A}^\mathsf{T} \mtx{\Pi}_{k, (i)} } }
\right\}
\end{align*}
is bounded by $\max\{\frac1d \matnorm{\mtx{A}}^2, \frac1d \matnorm{\mtx{A}}_{\rm F}^2\} = \frac1d \matnorm{\mtx{A}}_{\rm F}^2$.
Plugging these two values into Theorem~\ref{thm:intro-bernstein-vect}, we obtain
\begin{align*}
\Prob{\|\mtx{A}^\mathsf{T} \mtx{\Pi}_{(i)}\| > \sigma_A^0 + t} \leq 
\exp\paren{-\frac{t^2}{8 (\sigma_A^0)^2 + \frac{4R_A^0}{3} t }}.
\end{align*}
Furthermore, by applying a union bound argument we obtain
\begin{align*}
\Prob{\max_{i \in [d]} \|\mtx{A}^\mathsf{T} \mtx{\Pi}_{(i)}\| > \sigma_A^0 + t} \leq d \exp\paren{-\frac{t^2}{8 (\sigma_A^0)^2 + \frac{4R_A^0}{3} t }}.
\end{align*}
We can then substitute $\sqrt{8 (\sigma_A^0)^2 u} + \frac{4 R_A^0}{3}u$ for $t$ and have 
$$
\exp\paren{-\frac{t^2}{8 (\sigma_A^0)^2 + \frac{4 R_A^0}{3} t }} \leq 
\exp(-u).
$$
This will induce
$$
\Prob{\max_{i \in [d]} \|\mtx{A}^\mathsf{T} \mtx{\Pi}_{(i)}\| > R_A \defeq \frac43 u R_A^0 + (1+\sqrt{8u}) \cdot \sigma_A^0
} \leq d \cdot \exp(-u).
$$
We bound the right hand side of the display above by $\rho / 3$ to drop the bad case by setting $u \sim \log \frac{d}{\rho}$.
Thus with probability $1 - \frac\rho3$, $\|\mtx{A}^\mathsf{T} \mtx{\Pi}_{(i)}\|$'s are upper bounded by $R_A = \frac43 u R_A^0 + (1+\sqrt{8u}) \cdot \sigma_A^0$, 
and similarly with probability $1 - \frac\rho3$, $\|\mtx{B}^\mathsf{T} \mtx{\Pi}_{(i)}\|$'s are upper bounded by $R_B$.
($R_B$ can be determined by simply following the proof for $R_A$.)
$\|\mtx{X}_i\|$ are thus bounded by $R_A R_B + \frac1d \matnorm{\mtx{A}}\matnorm{\mtx{B}}$.
Our remaining task is to bound $\matnorm{\Expect\brkt {\mtx{X}_i \mtx{X}_i^\mathsf{T}}}$ and $\matnorm{\Expect\brkt {\mtx{X}_i^\mathsf{T} \mtx{X}_i}}$.

We start with $\matnorm{\Expect\brkt {\mtx{X}_i \mtx{X}_i^\mathsf{T}}}$. 
To give a finer analysis, we denote 
$$\mtx{\Pi}_{(i)}^\mathsf{T} = \frac{1}{\sqrt{md}} (Z_1, \cdots, Z_n),$$ 
and $Z_j$ can be further decomposed as $\frac{1}{\sqrt{p_j}} \sum_{k=1}^m Z_{jk}$, 
where $Z_{jk}$ is set corresponding to the $j$-th element in $\mtx{\Pi}_{k, (i)}$.
In summary, for a certain $i$, the $m$ random indicators in $\{Z_{jk}\}_{k=1}^m$ are i.i.d.\ and they would be $\pm 1$ with probability $\frac{p_j}{2}$ respectively, or be $0$ with probability $1-p_j$.
With the notation above, we can expand $\Expect\brkt {\mtx{X}_i \mtx{X}_i^\mathsf{T}}$ as
\begin{align}
\sum_{j_1, j_2, j_3, j_4 \in [n]} \Expect\brkt {\left(\frac{Z_{j_1} Z_{j_2}}{md} - \frac{\delta_{j_1 j_2}}{d}\right)\left(\frac{Z_{j_3} Z_{j_4}}{md} - \frac{\delta_{j_3 j_4}}{d}\right)} \mtx{A}_{(j_1)} \mtx{B}_{(j_2)}^\mathsf{T} \mtx{B}_{(j_3)} \mtx{A}_{(j_4)}^\mathsf{T}.
\end{align}

To simplify the summation, we make use of a key fact that if the index $j_1$ above is different than the other three indices, the corresponding summand will be $0$;
the similar results hold also for $j_2, j_3, j_4$.
Therefore we only need to consider the following four cases:
\begin{align*}
(1) j_1 = j_2 = j_3 = j_4, \quad (2) j_1 = j_2 \neq j_3 = j_4, \\
(3) j_1 = j_3 \neq j_2 = j_4, \quad (4) j_1 = j_4 \neq j_2 = j_3.
\end{align*}

Repeatedly using the fact that $\Expect\brkt {Z_{j_1} Z_{j_2}} = m \cdot \delta_{j_1 j_2}, 
\Expect\brkt {Z_{j_1}^4} = \frac{m}{p_i} + \binom{m}{2} \binom{4}{2}, \forall m \geq 2$, we can separately compute the sums in the four cases.
For the first one, we have 
\begin{align*}
&\Expect\brkt {\sum_{j_1 = j_2 = j_3 = j_4} \left(\frac{Z_{j_1}^2}{md} - \frac{1}{d}\right)^2 \mtx{A}_{(j_1)} \mtx{B}_{(j_1)}^\mathsf{T} \mtx{B}_{(j_1)} \mtx{A}_{(j_1)}^\mathsf{T}} \\
&\qquad\qquad\qquad = \frac{1}{d^2} \sum_{j_1=1}^n \left(\frac{1}{m p_{j_1}} + \frac{3(m-1)}{m} - 1\right) \mtx{A}_{(j_1)} \mtx{B}_{(j_1)}^\mathsf{T} \mtx{B}_{(j_1)} \mtx{A}_{(j_1)}^\mathsf{T};
\end{align*}
for the second case we have
\begin{align*}
&\Expect\brkt {\sum_{j_1 = j_2 \neq j_3 = j_4} \left(\frac{Z_{j_1}^2}{md} - \frac{1}{d}\right)\left(\frac{Z_{j_3}^2}{md} - \frac{1}{d}\right) \mtx{A}_{(j_1)} \mtx{B}_{(j_1)}^\mathsf{T} \mtx{B}_{(j_3)} \mtx{A}_{(j_3)}^\mathsf{T}} \\
&\qquad\qquad\qquad = -\frac{1}{m d^2} \sum_{j_1 \neq j_3} \mtx{A}_{(j_1)} \mtx{B}_{(j_1)}^\mathsf{T} \mtx{B}_{(j_3)} \mtx{A}_{(j_3)}^\mathsf{T};
\end{align*}
for the third case we have
\begin{align*}
\Expect\brkt {\sum_{j_1 = j_3 \neq j_2 = j_4} (\frac{Z_{j_1}Z_{j_2}}{md})^2 \mtx{A}_{(j_1)} \mtx{B}_{(j_2)}^\mathsf{T} \mtx{B}_{(j_1)} \mtx{A}_{(j_2)}^\mathsf{T}}
= \frac{m-1}{m d^2} \sum_{j_1 \neq j_2} \mtx{A}_{(j_1)} \mtx{B}_{(j_2)}^\mathsf{T} \mtx{B}_{(j_1)} \mtx{A}_{(j_2)}^\mathsf{T};
\end{align*}
for the fourth case we have
\begin{align*}
\Expect\brkt {\sum_{j_1 = j_4 \neq j_2 = j_3} (\frac{Z_{j_1}Z_{j_2}}{md})^2 \mtx{A}_{(j_1)} \mtx{B}_{(j_2)}^\mathsf{T} \mtx{B}_{(j_2)} \mtx{A}_{(j_1)}^\mathsf{T}}
= \frac{m-1}{m d^2} \sum_{j_1 \neq j_2} \mtx{A}_{(j_1)} \mtx{B}_{(j_2)}^\mathsf{T} \mtx{B}_{(j_2)} \mtx{A}_{(j_1)}^\mathsf{T}.
\end{align*}
A key observation is that $\Expect\brkt {(Z_{j_1}Z_{j_2})^2}$ can be computed by following the four-case division above, which is $m(m-1), \forall j_1 \neq j_2$.

Finally combining all the pieces together, we obtain
\begin{align*}
\Expect\brkt {\mtx{X}_i \mtx{X}_i^\mathsf{T}} = \frac{1}{m d^2} \sum_{j=1}^n \frac{\|\mtx{B}_{(j)}\|^2}{p_j} \mtx{A}_{(j)} \mtx{A}_{(j)}^\mathsf{T} + \frac{m-1}{m d^2} \matnorm{\mtx{B}}_{\rm F}^2 \mtx{A}^\mathsf{T} \mtx{A} + \frac{m-2}{m d^2} \mtx{A}^\mathsf{T} \mtx{B} \mtx{B}^\mathsf{T} \mtx{A},
\end{align*}
which implies
\begin{align*}
\matnorm{\Expect\brkt {\mtx{X}_i \mtx{X}_i^\mathsf{T}}} &\leq \frac{1}{d^2} \left(\frac{2s}{m\beta} \matnorm{\mtx{A}}^2 \matnorm{\mtx{B}}^2 + \frac{m-1}{m} \matnorm{\mtx{B}}_{\rm F}^2 \matnorm{\mtx{A}}^2 + \frac{m-2}{m} \matnorm{\mtx{A}^\mathsf{T} \mtx{B}}^2\right) \\
&\leq \frac{1}{d^2} \left(\frac{2s}{m \beta} + s + 1\right) \matnorm{\mtx{A}}^2 \matnorm{\mtx{B}}^2 ,
\end{align*}
For $\matnorm{\Expect\brkt {\mtx{X}_i^\mathsf{T} \mtx{X}_i}}$, we can similarly bound it as above by simply exchanging $\mtx{A}$ with $\mtx{B}$.

The final step of the proof involves applying Theorem~\ref{thm:intro-bernstein-rect}:
\begin{align*}
&\Prob{\matnorm{\mtx{A}^\mathsf{T} \mtx{\Pi}^\mathsf{T} \mtx{\Pi} \mtx{B} - \mtx{A}^\mathsf{T} \mtx{B}} > \varepsilon \matnorm{\mtx{A}} \matnorm{\mtx{B}}} \\
\leq& (p_A + p_B) \exp\left(\frac{-\varepsilon^2 \matnorm{\mtx{A}}^2 \matnorm{\mtx{B}}^2 /2}{\frac{1}{d} (\frac{2s}{m \beta} + s + 1) \matnorm{\mtx{A}}^2 \matnorm{\mtx{B}}^2 + \frac13 \varepsilon \matnorm{\mtx{A}} \matnorm{\mtx{B}} (R_A R_B + \frac1d \matnorm{\mtx{A}}\matnorm{\mtx{B}})}\right) \\
\leq& \frac{\rho}{3}.
\end{align*}
To make the inequality above hold, it suffices to ensure
\begin{align*}
\frac{1}{d} \paren{\frac{s}{m \beta} + s + 1} &\leq \frac{\varepsilon^2}{4\log \frac{3(p_A+p_B)}{\rho}} \\
R_A R_B + \frac1d \matnorm{\mtx{A}}\matnorm{\mtx{B}} &\leq \frac{32}{9} u^2 \frac{2s\matnorm{\mtx{A}}\matnorm{\mtx{B}}}{m d \beta} + 18 \frac{u s\matnorm{\mtx{A}}\matnorm{\mtx{B}}}{d} + \frac1d \matnorm{\mtx{A}}\matnorm{\mtx{B}} \\
&\leq \frac{3 \varepsilon \matnorm{\mtx{A}}\matnorm{\mtx{B}}}{4\log \frac{3(p_A+p_B)}{\rho}},
\end{align*}
which implies
\begin{align}
d &\gtrsim \frac{s}{\varepsilon} \log\frac{n}{\rho} \max\left\{\frac{1}{\varepsilon}, \log \frac{n}{\rho}\right\} \\
md &\gtrsim \frac{s}{\beta \varepsilon} \log\frac{n}{\rho} \max\left\{\frac{1}{\varepsilon}, \log^2 \frac{n}{\rho}\right\},
\end{align}
as Assumption~\ref{assum:size} holds.
\end{proof}

\section{Proof of Theorem~\ref{thm:tilde_K}}
\label{app:tilde_K}

\begin{proof}
The statement above can be divided into two parts, $\wt{\mtx{K}} \psdle \mtx{K}$ and $\mtx{K} \psdle \wt{\mtx{K}} + \lambda \mtx{I}$.
To prove them we first introduce some notations and auxiliary results.
Since $\mtx{K}$ is PSD, there exists a matrix $\mtx{B}$ satisfying $\mtx{B} \mtx{B}^\mathsf{T} = \mtx{K}$.
We further denote $\mtx{B}$'s SVD decomposition as $\mtx{B} = \mtx{U} \mtx{\Sigma}^{\frac12} \mtx{V}^\mathsf{T}$ (then $\mtx{K} = \mtx{U} \mtx{\Sigma} \mtx{U}^\mathsf{T}$),
where both $\mtx{U}$ and $\mtx{V}$ are $n$-by-$n$ orthogonal matrices.
Define $\bar{\mtx{\Sigma}} \defeq \mtx{\Sigma} + \lambda \mtx{I}, \mtx{\Psi} \defeq \mtx{U} \mtx{\Sigma}^{\frac12} \bar{\mtx{\Sigma}}^{-\frac12}$, implying $\mtx{K} (\mtx{K} + \lambda \mtx{I})^{-1} = \mtx{\Psi} \mtx{\Psi}^\mathsf{T}$.
With those notations, $\wt{\mtx{K}}$ can be rewritten as $\mtx{B} \mtx{B} \mtx{\Pi}^\mathsf{T} (\mtx{\Pi} \mtx{B} \mtx{B}^\mathsf{T} \mtx{\Pi}^\mathsf{T})^\dagger \mtx{\Pi} \mtx{B} \mtx{B}^\mathsf{T} = \mtx{B} \mtx{P_\Pi} \mtx{B}^\mathsf{T}$,
where $\mtx{P_\Pi}$ is the orthogonal projection matrix for the row space of $\mtx{\Pi} \mtx{B}$.
It is easy to check that $\mtx{K} - \wt{\mtx{K}} = \mtx{B} (\mtx{I} - \mtx{P_\Pi}) \mtx{B}^\mathsf{T}$.
Since $\mtx{I} - \mtx{P_\Pi}$ is an orthogonal projection matrix (which is PSD), 
we have $\mtx{K} - \wt{\mtx{K}} \psdge \mtx{0}$,
attaining the first conclusion that $\wt{\mtx{K}} \psdle \mtx{K}$.

For the second part, we utilize the following important identity:
\begin{align*}
\mtx{B}^\mathsf{T} \mtx{\Pi}^\mathsf{T} \mtx{\Pi} \mtx{B} - \mtx{B}^\mathsf{T} \mtx{B} 
&= \mtx{V} \bar{\mtx{\Sigma}}^{\frac12} \big( \bar{\mtx{\Sigma}}^{-\frac12} \mtx{V}^\mathsf{T} (\mtx{B}^\mathsf{T} \mtx{\Pi}^\mathsf{T} \mtx{\Pi} \mtx{B} - \mtx{B}^\mathsf{T} \mtx{B}) \mtx{V} \bar{\mtx{\Sigma}}^{-\frac12} \big) \bar{\mtx{\Sigma}}^{\frac12} \mtx{V}^\mathsf{T} \\
&= \mtx{V} \bar{\mtx{\Sigma}}^{\frac12} \big( \mtx{\Psi}^\mathsf{T} \mtx{\Pi}^\mathsf{T} \mtx{\Pi} \mtx{\Psi} - \mtx{\Psi}^\mathsf{T} \mtx{\Psi} \big) \bar{\mtx{\Sigma}}^{\frac12} \mtx{V}^\mathsf{T}.
\end{align*}
For $\mtx{\Psi}$, its squared Frobenius norm $\matnorm{\mtx{\Psi}}_{\rm F}^2 = d_{\rm stat}$, and $\matnorm{\mtx{\Psi}}^2 = \matnorm{\mtx{K} (\mtx{K} + \lambda \mtx{I})^{-1}} \geq 1/2$,
indicating that $\mtx{\Psi}$'s stable rank is at most $2 d_{\rm stat}$.
Taking $\varepsilon = \frac12$ and applying Theorem~\ref{thm:amm_sym}, we can conclude that given the conditions on $d$ and $md$ in the theorem, $\mtx{\Pi}$ satisfies $(\frac12, \rho)$-AMM property for $\mtx{\Psi}$.
Therefore it holds with probability $1-\rho$ that, 
$$\matnorm{\mtx{\Psi}^\mathsf{T} \mtx{\Pi}^\mathsf{T} \mtx{\Pi} \mtx{\Psi} - \mtx{\Psi}^\mathsf{T} \mtx{\Psi}} 
\leq \frac12 \matnorm{\mtx{\Psi}}^2
\leq \frac12.$$
From identity $\mtx{V} \bar{\mtx{\Sigma}}^{\frac12} \bar{\mtx{\Sigma}}^{\frac12} \mtx{V}^\mathsf{T} = \mtx{B}^\mathsf{T} \mtx{B} + {\lambda} \mtx{I}$, we obtain
\begin{align}
\frac12 \mtx{B}^\mathsf{T} \mtx{B} - \frac{\lambda}{2} \mtx{I} \psdle \mtx{B}^\mathsf{T} \mtx{\Pi}^\mathsf{T} \mtx{\Pi} \mtx{B} 
\psdle \frac32 \mtx{B}^\mathsf{T} \mtx{B} + \frac{\lambda}{2} \mtx{I},
\end{align}
which implies
\begin{align}
\label{eqn:BTB_bound}
\mtx{B}^\mathsf{T} \mtx{B} \psdle 2 \mtx{B}^\mathsf{T} \mtx{\Pi}^\mathsf{T} \mtx{\Pi} \mtx{B} + \lambda \mtx{I}.
\end{align}

Finally, we multiply two sides of equation~(\ref{eqn:BTB_bound}) by $(\mtx{I} - \mtx{P_\Pi})$ to obtain
\begin{align*}
(\mtx{I} - \mtx{P_\Pi}) \mtx{B}^\mathsf{T} \mtx{B} (\mtx{I} - \mtx{P_\Pi}) \psdle 2 \cdot \mtx{0} + \lambda (\mtx{I} - \mtx{P_\Pi}) \psdle \lambda \mtx{I},
\end{align*}
where the second inequality is due to the fact that $(\mtx{I} - \mtx{P_\Pi})$ is an orthogonal projection matrix.
The equation above implies 
$$\matnorm{(\mtx{I} - \mtx{P_\Pi}) \mtx{B}^\mathsf{T} \mtx{B} (\mtx{I} - \mtx{P_\Pi})} = \matnorm{\mtx{B} (\mtx{I} - \mtx{P_\Pi}) \mtx{B}^\mathsf{T}} \leq \lambda,$$ 
which completes the proof for the second conclusion $\mtx{K} \psdle \wt{\mtx{K}} + \lambda \mtx{I}$.
\end{proof}

The result above is interesting on its own, as it is somewhat counter-intuitive that directly seeking a small value of $\|\mtx{B}^\mathsf{T} \mtx{\Pi}^\mathsf{T} \mtx{\Pi} \mtx{B} - \mtx{B}^\mathsf{T} \mtx{B}\|$ is not an efficient way to obtain a good approximation $\wt{\mtx{K}}$. In this case, the projection dimension of $\mtx{\Pi}$ needs to be large enough to make the spectral norm discrepancy between $\mtx{B}^\mathsf{T} \mtx{\Pi} \mtx{\Pi^\mathsf{T}} \mtx{B}$ and $\mtx{B}^\mathsf{T} \mtx{B}$ small. Instead, we allow the discrepancy to be large in terms of $\mtx{B}^\mathsf{T} \mtx{\Pi}^\mathsf{T} \mtx{\Pi} \mtx{B}$ (note in the proof that $\mtx{B}^\mathsf{T} \mtx{B} - \mtx{B}^\mathsf{T} \mtx{\Pi}^\mathsf{T} \mtx{\Pi} \mtx{B} \psdle \mtx{B}^\mathsf{T} \mtx{\Pi}^\mathsf{T} \mtx{\Pi} \mtx{B} + \lambda \mtx{I}$), which will be eliminated in $\mtx{K} - \wt{\mtx{K}}$ thanks to the orthogonal projection matrix $(\mtx{I} - \mtx{P_\Pi})$.

In this proof, we have borrowed some concepts and proof techniques from the literature on KRR. The statistical dimension $d_{\rm stat} = \Tr\paren{\mtx{K} (\mtx{K} + \lambda \mtx{I})^{-1}}$ mentioned in the theorem is closely related to the stable rank of $\mtx{\Psi}$ (recall $\mtx{K} (\mtx{K} + \lambda \mtx{I})^{-1} = \mtx{\Psi} \mtx{\Psi}^\mathsf{T}$), since in kernel methods, $\lambda$ is usually much smaller than $\matnorm{\mtx{K}}$, and thus $\matnorm{\mtx{\Psi} \mtx{\Psi}^\mathsf{T}}$ is close to $1$. This fact implies that the statistical dimension $d_{\rm stat}$, which is similar to the stable rank (we note $s = \matnorm{\mtx{\Psi}}_{\rm F}^2 / \matnorm{\mtx{\Psi}}^2 \approx d_{\rm stat}$), reflects the minimal number of sub-samples needed to recover the original matrix $\mtx{K}$. These concepts lead to a better understanding of the seemingly unnatural parameter $\lambda$. Here, the constant $\lambda > 0$ plays a role as a tuning parameter balancing computational complexity and numerical accuracy. A small $\lambda$ leads to a more accurate approximation $\wt{\mtx{K}}$ but requires more sub-samples in $\mtx{\Pi}$. Note that when $\lambda \to 0$, the statistical dimension $d_{\rm stat} = \Tr \big( \mtx{K} (\mtx{K} + \lambda \mtx{I})^{-1} \big) \to n$, and we may need almost all the samples to guarantee an accurate approximation.
The above intuition can also be explained by the following derivation:
\begin{align*}
\mtx{B}^\mathsf{T} \mtx{\Pi} \mtx{\Pi^\mathsf{T}} \mtx{B} - \mtx{B}^\mathsf{T} \mtx{B} 
= \mtx{V} \mtx{\Sigma}^{\frac12} \big( \mtx{U}^\mathsf{T} \mtx{\Pi} \mtx{\Pi}^\mathsf{T} \mtx{U} - \mtx{U}^\mathsf{T} \mtx{U} \big) \mtx{\Sigma}^{\frac12} \mtx{V}^\mathsf{T} 
\psdle \matnorm{\mtx{U}^\mathsf{T} \mtx{\Pi} \mtx{\Pi}^\mathsf{T} \mtx{U} - \mtx{I}} \mtx{B}^\mathsf{T} \mtx{B}.
\end{align*}
To ensure $\matnorm{\mtx{U}^\mathsf{T} \mtx{\Pi} \mtx{\Pi}^\mathsf{T} \mtx{U} - \mtx{I}} \leq \frac12$, the dimension of $\mtx{\Pi}$ needs to be at least $\text{rank}(\mtx{B})$. In fact, according to the Eckart-Young Theorem, even truncated SVD, which provides the best low-rank approximation, will suffer a spectral loss of $1$. In kernel ridge regression and many other applications, the rank of $\mtx{B}$ is typically $n$ (sample size), so the dimension of $\mtx{\Pi}$ needs to be at least $n$.

\section{Extended discussion about the proof of Theorem~\ref{thm:rsvd}}
\label{sec:grproof}

In this section, we investigate the effectiveness of our proposed \algoname{} for randomized SVD.
Specifically, we take the estimator in generalized linear regression as an intermediate object, whose error happens to upper bound the one in randomized SVD.
In Supplement~\ref{sec:intro-gr}, we show our method can also apply to generalized linear regression and attain desired approximation.
With the result on generalized linear regression, we lay out the brief proof of Theorem~\ref{thm:rsvd} in Supplement~\ref{sec:proof-rsvd}.

\subsection{Introduction to generalized linear regression}
\label{sec:intro-gr}

As an extension of linear regression, generalized linear regression aims to project a  ``response'' matrix $\mtx{B}$, rather than a vector, onto the column space of a given design matrix $\mtx{A}$ with left singular vectors $\mtx{U_A}$. 
Based on the Eckart--Young Theorem, the optimal approximation that minimizes both the Frobenius and spectral norms is $\mtx{A} \mtx{X} = \mtx{P_A} \mtx{B}$ with $\mtx{X} \defeq (\mtx{A}^\mathsf{T} \mtx{A})^\dagger \mtx{A}^\mathsf{T} \mtx{B}$, where $\mtx{P_A}$ is the orthogonal projection matrix for the column space of $\mtx{A}$. A natural approximation method is to replace $\mtx{A}$ and $\mtx{B}$ with their sketched approximations $\mtx{\Pi}\mtx{A}$ and $\mtx{\Pi} \mtx{B}$, respectively, in order to efficiently compute $\wt{\mtx{X}} = (\mtx{A}^\mathsf{T} \mtx{\Pi}^\mathsf{T} \mtx{\Pi} \mtx{A})^\dagger \mtx{A}^\mathsf{T} \mtx{\Pi}^\mathsf{T} \mtx{\Pi} \mtx{B}$. 
\citet[Theorem~3]{DBLP:conf/icalp/CohenNW16} showed that the quantity $\matnorm{\mtx{A} \wt{\mtx{X}} - \mtx{B}}$ is small when $\matnorm{\mtx{U_A}^\mathsf{T} \mtx{\Pi}^\mathsf{T} \mtx{\Pi} \mtx{U_A} - \mtx{I}_{n}} \leq 1/2$ and $\matnorm{\mtx{U_A}^\mathsf{T} \mtx{\Pi}^\mathsf{T} \mtx{\Pi} (\mtx{I} - \mtx{P_A}) \mtx{B}} \ll 1$. 
In this case, the specification of the sampling probabilities required in Theorem~\ref{thm:amm_asym} will be formulated as 
\begin{align}
\label{eqn:lvg-scores}
p_j \geq \beta \frac{\max\{\|[\mtx{U_A}]_{(j)}\|^2, \|[(\mtx{I} - \mtx{P_A}) \mtx{B}]_{(j)}\|^2 / \matnorm{(\mtx{I} - \mtx{P_A}) \mtx{B}}^2\}}{\sum_{j'=1}^{n} \max\{\|[\mtx{U_A}]_{(j')}\|^2, \|[(\mtx{I} - \mtx{P_A}) \mtx{B}]_{(j')}\|^2 / \matnorm{(\mtx{I} - \mtx{P_A}) \mtx{B}}^2\}}, \quad \forall j = 1,\dots,n.
\end{align}
Here, $\{\|[\mtx{U_A}]_{(j)}\|^2 / p_A\}_{j=1}^{n}$ is the so-called statistical leverage scores~\citep{drineas2006fast} of matrix $\mtx A$.

In the following theorem, we respectively apply Theorems~\ref{thm:amm_sym} and~\ref{thm:amm_asym} to provide error and dimension bounds based on their proof. The proof is deferred to Supplement~\ref{sec:proof-gr}.
Specifically, we discuss the performance of uniform sampling, which is commonly used in practice due to the hardness of computing statistical leverage scores~\citep{chen2021skyformer}.
\begin{theorem}
\label{thm:gr}
For matrices $\mtx{A} \in \mb R^{n\times p_A}, \mtx{B} \in \mb R^{n\times p_B}$,
we let $s$ be the maximum of $p_A$ and $(\mtx{I} - \mtx{P_A}) \mtx{B}$'s stable rank.
Suppose $\mtx{\Pi}$ is an $\paren{m, d, \{p_j\}_{j=1}^{n}}$-\algoname{} matrix where $\{p_j\}_{j=1}^{n}$ satisfy equation~\eqref{eqn:lvg-scores}. 
For constant $\varepsilon > 0, \rho < \frac12$, there exists an absolute constant $C$ such that if
\begin{align*}
d \geq C \frac{s}{\sqrt \varepsilon} \log\frac{n}{\rho} \max\left\{\frac{1}{\sqrt \varepsilon}, \log \frac{n}{\rho}\right\}, \quad
md \geq C \frac{s^2}{p_A \beta \sqrt \varepsilon} \log\frac{n}{\rho} \max\left\{\frac{1}{\sqrt \varepsilon}, \log^2 \frac{n}{\rho}\right\},
\end{align*}
then $\matnorm{\mtx{A} \wt{\mtx{X}} - \mtx{B}} \leq (1+\varepsilon) \matnorm{(\mtx{I} - \mtx{P_A}) \mtx{B}}^2$ with probability $1-\rho$.
Furthermore, if one use uniform sampling and $p_j = \frac1n, \forall j \in [n]$, then the dimension condition is
\begin{align*}
d \gtrsim \frac{s}{\sqrt \varepsilon} \log\frac{n}{\rho} \max\left\{\frac{1}{\sqrt \varepsilon}, \log \frac{n}{\rho}\right\}, \quad
md \gtrsim \frac{n s}{p_A \sqrt \varepsilon} \log\frac{n}{\rho} \max\left\{\frac{1}{\sqrt \varepsilon}, \log^2 \frac{n}{\rho}\right\},
\end{align*}
so as to attain $\matnorm{\mtx{A} \wt{\mtx{X}} - \mtx{B}} \leq (1+\varepsilon) \matnorm{(\mtx{I} - \mtx{P_A}) \mtx{B}}^2$ with probability $1-\rho$.
\end{theorem}

\noindent \textbf{Complexity analysis:} 
As Theorem~\ref{thm:amm_asym} suggests, the projection dimension $d$ of our method is roughly the same as Gaussian sketching, and we can safely assume that the projection dimension is the same for all other data-oblivious methods. For the sub-sampling probabilities, since pre-computing matrices $\mtx{U_A}$ and $(\mtx{I} - \mtx{P_A}) \mtx{B}$ is difficult, we simply use the uniform sub-sampling scheme. In the worst case, $md = \wt{\m O}(n)$ ($\widetilde{\mathcal{O}}$ stands for $\mathcal O$ modulo poly-log terms), and the complexity of performing generalized linear regression becomes $\wt{\m O}(n(p_A + p_B))$. Gaussian sketching takes $\m O(n(p_A + p_B)d)$ time, SRHT takes $\m O(n(p_A + p_B) \log d) = \wt{\m O}(n(p_A + p_B))$ time, sparse OSE takes $\wt{\m O}(n(p_A + p_B))$ time, and sub-sampling sketching method takes $\wt{\m O}(n p_A p_B)$ time as it requires $md$ as the projection dimension. We remark that the intrinsic complexity of generalized linear regression is high, but in some applications, such as randomized SVD discussed below, we can instead compress $p_A$ rather than $n$.

\subsection{Proof of Theorem~\ref{thm:rsvd}}
\label{sec:proof-rsvd}

The proof of Theorem~\ref{thm:rsvd} is direct.
We know the result $\wt{\mtx{A}}_k$ obtained by randomized SVD is the \emph{best} possible rank-$k$ approximation (w.r.t.\ the spectral norm) to $\mtx{A}$ that lies within the row space of ${\mtx{\Pi}} \mtx{A}$~\cite[Theorem~4]{DBLP:conf/icalp/CohenNW16}. 
In this case, we directly have $\matnorm{\wt{\mtx{A}}_k - \mtx{A}}^2 
\leq \matnorm{\mtx{A}_k \wt{\mtx{X}} - \mtx{A}}^2$,
and our task reduces to proving $\matnorm{\mtx{A}_k \wt{\mtx{X}} - \mtx{A}}^2
\leq (1+\varepsilon) \matnorm{\bar{\mtx A}_k}^2$.
Realizing the correspondence that the $\mtx U_k$ and $\bar{\mtx A}_k$ in Theorem~\ref{thm:rsvd} are just the $\mtx{U_A}$ and $(\mtx{I} - \mtx{P_A}) \mtx{B}$ in Theorem~\ref{thm:gr},
we can then prove $\matnorm{\mtx{A}_k \wt{\mtx{X}} - \mtx{A}}^2
\leq (1+\varepsilon) \matnorm{\bar{\mtx A}_k}^2$ will hold with probability at least $1-\rho$.
The claim in Theorem~\ref{thm:rsvd} is then attained as well.

\subsection{Technical proof of Theorem~\ref{thm:gr}}
\label{sec:proof-gr}

As derived by \citet[Theorem~3]{DBLP:conf/icalp/CohenNW16}, 
\begin{align}
\label{eqn:gr}
\matnorm{\mtx{A} \wt{\mtx{X}} - \mtx{B}} 
\leq \matnorm{(\mtx{U_A}^\mathsf{T} \mtx{\Pi}^\mathsf{T} \mtx{\Pi} \mtx{U_A})^\dagger}^2 \matnorm{\mtx{U_A}^\mathsf{T} \mtx{\Pi}^\mathsf{T} \mtx{\Pi} (\mtx{I} - \mtx{P_A}) \mtx{B}}^2 + \matnorm{(\mtx{I} - \mtx{P_A}) \mtx{B}}^2.
\end{align}
Our remaining task is to decide the thresholds for $d$ and $md$ so that \begin{align}
\label{eqn:gr-condition}
\matnorm{\mtx{U_A}^\mathsf{T} \mtx{\Pi}^\mathsf{T} \mtx{\Pi} \mtx{U_A} - \mtx{I}_{p_A}} < \frac12 
~~\text{and}~~ \matnorm{\mtx{U_A}^\mathsf{T} \mtx{\Pi}^\mathsf{T} \mtx{\Pi} (\mtx{I} - \mtx{P_A}) \mtx{B}} < \sqrt{\frac\varepsilon4}
\end{align} 
with probability $1 - \frac\rho2$. 
To simplify the notation, we denote $\bar{\mtx{B}} \defeq (\mtx{I} - \mtx{P_A}) \mtx{B}$.

We first apply Theorem~\ref{thm:amm_sym} to address $\matnorm{\mtx{U_A}^\mathsf{T} \mtx{\Pi}^\mathsf{T} \mtx{\Pi} \mtx{U_A} - \mtx{I}_{p_A}}$.
Notably, the reciprocal of the quality coefficient in this case will be bounded by
$$
\max_j \set{\frac{1}{p_j} \frac{\|[\mtx{U_A}]_{(j)}\|^2}{p_A} } 
\leq \frac{1}{\beta} \sum_{j'=1}^{n} \max\{\|[\mtx{U_A}]_{(j')}\|^2, \|\bar{\mtx{B}}_{(j')}\|^2 / \matnorm{\bar{\mtx{B}}}^2\}
\leq \frac{1}{\beta} \frac{2s}{p_A}.
$$
Theorem~\ref{thm:amm_sym} then gives
\begin{align*}
d \geq C 2s \log\frac{n}{\rho} \max\left\{2, \log \frac{n}{\rho}\right\}, \quad
md \geq C \frac{4 s^2}{p_A \beta} \log\frac{n}{\rho} \max\left\{2, \log^2 \frac{n}{\rho}\right\}. 
\end{align*}
We then apply Theorem~\ref{thm:amm_asym} to the event $\matnorm{\mtx{U_A}^\mathsf{T} \mtx{\Pi}^\mathsf{T} \mtx{\Pi} \bar{\mtx{B}}} < \sqrt{\frac\varepsilon4}$, which gives
\begin{align*}
d \geq C \frac{2s}{\sqrt \varepsilon} \log\frac{n}{\rho} \max\left\{\frac{2}{\sqrt \varepsilon}, \log \frac{n}{\rho}\right\}, \quad
md \geq C \frac{2s}{\beta \sqrt \varepsilon} \log\frac{n}{\rho} \max\left\{\frac{2}{\sqrt \varepsilon}, \log^2 \frac{n}{\rho}\right\}. 
\end{align*}
In summary, we have the following specification to satisfy both conditions on $d, md$:
\begin{align*}
d \geq C \frac{s}{\sqrt \varepsilon} \log\frac{n}{\rho} \max\left\{\frac{1}{\sqrt \varepsilon}, \log \frac{n}{\rho}\right\}, \quad
md \geq C \frac{s^2}{p_A \beta \sqrt \varepsilon} \log\frac{n}{\rho} \max\left\{\frac{1}{\sqrt \varepsilon}, \log^2 \frac{n}{\rho}\right\}. 
\end{align*}
In that case, $\matnorm{(\mtx{U_A}^\mathsf{T} \mtx{\Pi}^\mathsf{T} \mtx{\Pi} \mtx{U_A})^\dagger} \leq 2$ and equation~(\ref{eqn:gr}) is bounded by 
$$
(1+2^2 \frac\varepsilon4) \matnorm{\bar{\mtx{B}}}^2 = (1+\varepsilon) \matnorm{\bar{\mtx{B}}}^2.
$$
We then turn to prove the results for the special case where $p_j = \frac1n, \forall j$.
Again, we need to bound the quality coefficient of uniform sampling to control the total number of sub-samples.
For $\matnorm{\mtx{U_A}^\mathsf{T} \mtx{\Pi}^\mathsf{T} \mtx{\Pi} \mtx{U_A} - \mtx{I}_{p_A}}$, 
in which the reciprocal of the quality coefficient should be bounded by $\max_j \frac{\|(\mtx{U_A})_{(j)}\|}{p_j p_A} \leq \frac{n}{p_A}$;
in applying Theorem~\ref{thm:amm_asym} to $\matnorm{\mtx{U_A}^\mathsf{T} \mtx{\Pi}^\mathsf{T} \mtx{\Pi} \bar{\mtx{B}}}$, 
we utilize the facts that 
\begin{align*}
\max\{\|(\mtx{U_A})_{(j)}\|^2, \|\bar{\mtx{B}}_{(j)}\|^2 / \matnorm{\bar{\mtx{B}}}^2\} &\leq 1,
\end{align*}
and
\begin{align*}
s &= \max\set{\sum_j \|(\mtx{U_A})_{(j)}\|^2, \sum_i \|\bar{\mtx{B}}_{(j)}\|^2 / \matnorm{\bar{\mtx{B}}}^2} \\
&\leq \sum_j \max\set{\|(\mtx{U_A})_{(j)}\|^2, \|\bar{\mtx{B}}_{(j)}\|^2 / \matnorm{\bar{\mtx{B}}}^2},
\end{align*}
and derive the reciprocal of the quality coefficient should be bounded by
\begin{align*}
\max_j \frac{1}{p_j} \frac{\max\{\|(\mtx{U_A})_{(i)}\|^2, \|\bar{\mtx{B}}_{(i)}\|^2 / \matnorm{\bar{\mtx{B}}}^2\}}{\sum_{j=1}^{n} \max\{\|(\mtx{U_A})_{(j)}\|^2, \|\bar{\mtx{B}}_{(j)}\|^2 / \matnorm{\bar{\mtx{B}}}^2\}}
\leq n \frac{1}{s}.
\end{align*}

Combining all the pieces above, we validate that with 
\begin{align}
d \gtrsim \frac{s}{\sqrt \varepsilon} \log\frac{n}{\rho} \max\left\{\frac{1}{\sqrt \varepsilon}, \log \frac{n}{\rho}\right\}, \quad
md \gtrsim \frac{n s}{p_A \sqrt \varepsilon} \log\frac{n}{\rho} \max\left\{\frac{1}{\sqrt \varepsilon}, \log^2 \frac{n}{\rho}\right\},
\end{align}
by Theorems~\ref{thm:amm_sym} and~\ref{thm:amm_asym} our sketching matrix $\mtx{\Pi}$ can make both conditions in \eqref{eqn:gr-condition}
hold with probability $1-\rho/2$.
The claim is then analogously attained.

\section{Extended discussion about $K$-satisfiability}

We will first introduce some preliminary knowledge of \emph{reproducing kernel Hilbert space} (RKHS in shorthand) kernels as well as technical assumptions to control the eigenvalue behavior of the kernel function and the empirical kernel matrix; 
we then compare the magnitude of statistical dimension $d_{\rm stat}$ with $d_\delta$, and provide the complete proof of Theorem~\ref{thm:k-satis} to show $K$-satisfiability.

\subsection{Assumptions on RKHS Kernels}

\cite{yang2017randomized} guaranteed that for a $K$-satisfiable sketching matrix $\mtx \Pi$, 
with high probability the approximation error $\|\hat f_S - \hat f_n\|_n^2$ would be bounded by $\lambda + \frac{d_\lambda}{n}$.
The result is powerful, while it relies on some assumptions on the kernel function and the sketching matrix used.
In this subsection, we first provide the necessary introduction to the eigendecomposition of the kernel function $\m K$, so as to ease the explanation of the two assumptions summarized by \cite{liu2018nonparametric}.

An RKHS kernel function $\m K$, by Mercer’s theorem, has the following spectral expansion:
\begin{align*}
\m K(x, x') = \sum_{i=1}^\infty \mu_i \phi_i(x) \phi_i(x'), \quad \forall x, x' \in \m X,
\end{align*}
where $\mu_1 \geq \mu_2 \geq \cdots \geq 0$ are denoted as the eigenvalues of $\m K$,
and $\{\phi_i\}_{i=1}^\infty$ actually form a basis in $L^2(\m X)$, i.e.
\begin{align*}
\dotp{\phi_i}{\phi_j}_{L^2(\m X)} = \delta_{ij}, \quad \dotp{\phi_i}{\phi_j}_{\mb H} = \delta_{ij} / \mu_i.
\end{align*}

The eigenvalues $\{\mu_i\}_{i=1}^\infty$ of the kernel function $\m K$ are closely related to the eigenvalues $\{\sigma_i\}_{i=1}^n$ of the re-scaled empirical kernel matrix $\frac{1}{n} \mtx K$.
The eigenvalue pair $(\mu_i, \sigma_i)$ of the same index $i$ would roughly have the same magnitude, and more details can be found in the work \citep{braun2006accurate}.
We can then state the two assumptions made by \cite{liu2018nonparametric} as follows.
\begin{assumption}
\label{Assu:1}
Let $c_K \defeq \sup_{i \geq 1} \|\phi_i\|_{\sup} < \infty$, and $\sup_{k \geq 1} \frac{\sum_{i=k+1}^\infty \mu_i}{k \mu_k} < \infty$.
\end{assumption}
Assumption~\ref{Assu:1} requires the kernel function to have a fast eigenvalue decay rate. 
Fortunately, many common kernel functions satisfy the assumption.
For example, the \matern kernel with smoothness parameter $\nu$ has a decay rate as $\mu_i \asymp i^{-\frac{2\nu + p_X}{p_X}}$ \citep{bach2017equivalence}, and we can check the rate of this type will satisfy Assumption~\ref{Assu:1};
the Gaussian kernels have an even faster decay rate $\mu_i \asymp \exp(-\gamma \cdot i^c)$, (where $\gamma, c$ are some constants) and also satisfy the assumption.

The next assumption has further requirements on the eigenvalue sequence:
\begin{assumption}
\label{Assu:2}
Let $s_\lambda \defeq \min\{i: \mu_i \leq \lambda\} - 1$.
$s_\lambda$ diverges as $\lambda \to 0$.
\end{assumption}
With this assumption, kernels will have a sequence of positive eigenvalues converging to $0$.
Most infinitely dimensional kernels satisfy this assumption, including the two examples above, \matern and Gaussian kernels.

\subsection{Comparing statistical dimension $d_{\rm stat}$ with $d_\delta$}
\label{app:d_stat_delta}

To analyze the connection between $d_{\rm stat}$ and $d_\delta$, 
we set $\lambda = c \delta^2$ (i.e., $\lambda = \Omega(\delta^2)$) for a constant $c \geq 2$. 
We first lower bound $d_{\rm stat}$ by $d_\delta$:
\begin{align}
\label{eqn:lb4dstat}
d_{\rm stat} = \sum_{j=1}^n \frac{\sigma_j}{\sigma_j + c \delta^2}
> \frac1c \sum_{j=1}^n \frac{\sigma_j}{\sigma_j + \delta^2}
\geq \frac1c \sum_{j=1}^{d_\delta} \frac{\sigma_j}{\sigma_j + \delta^2}
> \frac{d_\delta}{2c}.
\end{align}

For the other side that $d_{\rm stat}$ is upper bounded by a constant multiple of $d_\delta$, 
we can again separate the terms in $d_{\rm stat} = \sum_{j=1}^n \frac{\sigma_j}{\sigma_j + c \delta^2}$ by the cutoff $\delta^2$:
\begin{align}
\label{eqn:ub4dstat}
d_{\rm stat} = \sum_{j=1}^n \frac{\sigma_j}{\sigma_j + c \delta^2}
< \sum_{j=1}^n \frac{\sigma_j}{\sigma_j + \delta^2}
< d_\delta + \sum_{j=d_\delta+1}^{n} \frac{\sigma_j}{\sigma_j + \delta^2},
\end{align}
where without loss of generality we assume $n > d_\delta$, since otherwise $d_{\rm stat} = \sum_{j=1}^n \frac{\sigma_j}{\sigma_j + c \delta^2}$ is upper bounded by $n \leq d_\delta$ and we directly obtain the other side of inequality.
Further bounding equation~\eqref{eqn:ub4dstat}, we have:
\begin{align}
\label{eqn:ub4dstat2}
d_{\rm stat} < d_\delta + \sum_{j=d_\delta+1}^{n} \frac{\sigma_j}{\sigma_j + \delta^2} 
\leq d_\delta + \frac{1}{\delta^2} \sum_{j=d_\delta+1}^{n} \sigma_j
\leq d_\delta + \frac{1}{\delta^2} c' d_\delta \delta^2
= (1+c') d_\delta,
\end{align}
in which the last inequality is induced by the fast eigenvalue decay rate in Assumption~\ref{Assu:1} and will hold with high probability \citep[Lemma~3.1 and Lemma~S.2 (a)]{liu2018nonparametric}.

Equations (\ref{eqn:lb4dstat}) and (\ref{eqn:ub4dstat2}) together indicate the statistical dimension for using $\lambda$ is of the same order as $d_{\delta}$.
Furthermore, for $\matnorm{\mtx{\Psi}}_{\rm F}^2 = \sum_{j=d_\delta+1}^{n} \frac{\sigma_j}{\sigma_j + \delta^2}$, we have
\begin{align}
d_{\rm stat} < \matnorm{\mtx{\Psi}}_{\rm F}^2 = \sum_{j=d_\delta+1}^{n} \frac{\sigma_j}{\sigma_j + \delta^2} < c \, d_{\rm stat},
\end{align}
through combining the derivation in equations (\ref{eqn:lb4dstat}) and (\ref{eqn:ub4dstat2}). 
That implies $\matnorm{\mtx{\Psi}}_{\rm F}^2$ is of the same order as $d_{\delta}$ as well.

\subsection{Proof of Theorem~\ref{thm:k-satis}}
\label{app:k-satis}

This is a complete version of the proof to show $K$-satisfiability.
\begin{align*}
\matnorm{\mtx{U}_1^\mathsf{T} \mtx{\Pi}^\mathsf{T} \mtx{\Pi} \mtx{U}_1 - \mtx I_{d_{\delta}}} \leq 1/2, \quad \mathrm{and} \quad
\matnorm{\mtx{\Pi} \mtx{U}_2 \mtx{\Sigma}_2^{\frac{1}{2}}} \leq c \delta.
\end{align*}
\begin{proof}
For the first condition $\matnorm{\mtx{U}_1^\mathsf{T} \mtx{\Pi}^\mathsf{T} \mtx{\Pi} \mtx{U}_1 - \mtx I_{d_{\delta}}} \leq 1/2$, we notice that it suffices to prove $(\frac12, \frac13\rho)$-AMM property for $\mtx{U}_{1}$, whose stable rank is exactly $d_{\delta}$.
For the the quality coefficient, utilizing the fact that $\frac{\sigma_j}{\sigma_j + \delta^2} \geq \frac12, \forall j = 1,\dots,d_\delta$,
we can bound $\frac{\|(\mtx{U}_{1})_{(j)}\|^2}{p_j}$ from above by $\frac{2 \matnorm{\mtx{\Psi}}_{\rm F}^2}{\beta} \sim \frac{d_\delta}{\beta}$, 
utilizing the analysis in Section~\ref{app:d_stat_delta}.
Combining the pieces above into the proof of Theorem~\ref{thm:amm_sym}, we obtain that the first condition in Definition~\ref{def:k_satis} holds with probability $1-\frac{\rho}{3}$, under the following conditions:
\begin{align}
\label{eqn:k-satis-cond}
d \geq C_6 d_\delta \log^2\frac{n}{\rho}, \quad
md \geq C_6 \frac{d_\delta}{\beta} \log^3\frac{n}{\rho}.
\end{align}
We remark the requirement on $md$ will be strengthened in the following derivation for the second condition of $K$-satisfiability.

For the second condition $\matnorm{\mtx{\Pi} \mtx{U}_2 \mtx{\Sigma}_2^{\frac{1}{2}}} \leq c \delta$, 
we can utilize matrix Chernoff bound in Theorem~\ref{thm:intro-chernoff}.
Following the notations in Theorem~\ref{thm:intro-chernoff}, we denote $\mu_{\max} \defeq \matnorm{\mtx{\Sigma}_2^{\frac{1}{2}} \mtx{U}_2^\mathsf{T} \mtx{U}_2 \mtx{\Sigma}_2^{\frac{1}{2}}} = \sigma_{d_\delta + 1}$;
we further pick $\eta$ satisfying $(1 + \eta) \mu_{\max} = 2 \delta^2$, which implies $\eta > 1$. 
In this case,
\begin{align}
\Prob{ \matnorm{\mtx{\Sigma}_2^{\frac{1}{2}} \mtx{U}_2^\mathsf{T} \mtx{\Pi}^\mathsf{T} \mtx{\Pi} \mtx{U}_2 \mtx{\Sigma}_2^{\frac{1}{2}}} \geq (1 + \eta) \mu_{\max} = 2 \delta^2}
	\leq (n - d_\delta) \cdot \left[ \frac{\econst^{\eta}}{(1+\eta)^{1 + \eta}} \right]^{\mu_{\max}/R},   
\end{align}
where $R = \max_{i \in [d]} \matnorm{\mtx{\Sigma}_2^{\frac{1}{2}} \mtx{U}_2^\mathsf{T} \mtx{\Pi}_{(i)} \mtx{\Pi}_{(i)}^\mathsf{T} \mtx{U}_2 \mtx{\Sigma}_2^{\frac{1}{2}} }$.
To further bound the above right-hand-side expression and relate it to $\delta$,
we can utilize Lemma~\ref{lem:chernoff_bd} and attain:
\begin{align*}
\Prob{ \matnorm{\mtx{\Sigma}_2^{\frac{1}{2}} \mtx{U}_2^\mathsf{T} \mtx{\Pi}^\mathsf{T} \mtx{\Pi} \mtx{U}_2 \mtx{\Sigma}_2^{\frac{1}{2}}} \geq 2 \delta^2}
	\leq n \cdot \left[ \econst^{-\frac{1}{3}\eta} \right]^{\frac{2\delta^2}{1 + \eta}/R}
\leq n \cdot \exp\paren{-\frac{2\delta^2}{3R}} \leq \rho/3.
\end{align*}
To make the second condition in $K$-satisfiability hold, we only need to validate
\begin{align}
\label{eqn:sec_cond_R}
R / \delta^2 \lesssim 1 / \log \frac{n}{\rho}.
\end{align}

In the following standard derivation, we will specify the scale of $R$ to complete the proof.
We first denote $\bar{\mtx{\Sigma}}_2 \defeq \mtx{\Sigma}_2 + \delta^2 \mtx{I}_{n-d_{\delta}}$, 
$\mtx{\Psi}_2 = \mtx{U}_2 \mtx{\Sigma}_2^{\frac{1}{2}} \bar{\mtx{\Sigma}}_2^{-\frac12}$, and rewrite $\mtx{\Pi} \mtx{U}_2 \mtx{\Sigma}_2^{\frac{1}{2}} = \mtx{\Pi} \mtx{\Psi}_2 \bar{\mtx{\Sigma}}_2^{\frac12}$, so as to leverage the given conditions on sampling probabilities.
Similar to the proof in Supplement~\ref{sec:amm_asym}, we recall
\begin{align*}
\mtx{\Pi}_{(i)} = \sum_{k=1}^m \frac{1}{\sqrt{m}} \mtx{\Pi}_{k, (i)}, \quad \forall i \in [d].
\end{align*}
For $\frac{1}{\sqrt{m}} \bar{\mtx{\Sigma}}_2^{\frac12} \mtx{\Psi}_2^\mathsf{T} \mtx{\Pi}_{k, (i)}$, we can bound its operator norm as
\begin{align*}
\norm{\frac{1}{\sqrt{m}} \bar{\mtx{\Sigma}}_2^{\frac12} \mtx{\Psi}_2^\mathsf{T} \mtx{\Pi}_{k, (i)} } 
&\leq \max_{j} \sqrt{\frac{2 \delta^2}{m d}} \frac{\|\mtx{\Psi}_{2, {(j)}}\|}{\sqrt{p_j}}
\leq \|\mtx{\Psi}_{2, {(j)}}\| \sqrt{\frac{2 \delta^2 \matnorm{\mtx{\Psi}}_{\rm F}^2}{m d \beta \|\mtx{\Psi}_{(j)}\|^2}} \\
&\defeq R_2 \sim \sqrt{\frac{\delta^2 d_\delta}{m d} }.
\end{align*}
For the ``variance'' term $\sigma_2^2$, it is defined as
\begin{align*}
\max\left\{ 
	\matnorm{ \frac1m \sum_{k=1}^m \Expect\brkt{ 
 \mtx{\Sigma}_2^{\frac{1}{2}} \mtx{U}_2^\mathsf{T} \mtx{\Pi}_{k, (i)} \mtx{\Pi}_{k, (i)}^\mathsf{T} \mtx{U}_2 \mtx{\Sigma}_2^{\frac{1}{2}}  } }, \
	\matnorm{ \frac1m \sum_{k=1}^m \Expect\brkt{\mtx{\Pi}_{k, (i)}^\mathsf{T} \mtx{U}_2 \mtx{\Sigma}_2 \mtx{U}_2^\mathsf{T} \mtx{\Pi}_{k, (i)} } }
\right\}
\end{align*}
and is bounded by $\max\{\frac1d \matnorm{\mtx{U}_2 \mtx{\Sigma}_2^{\frac{1}{2}}}^2, \frac1d \matnorm{\mtx{U}_2 \mtx{\Sigma}_2^{\frac{1}{2}}}_{\rm F}^2\} = \frac1d \matnorm{\mtx{U}_2 \mtx{\Sigma}_2^{\frac{1}{2}}}_{\rm F}^2 \sim \frac{\delta^2 d_\delta}{d}$.
By further applying a union bound argument to Theorem~\ref{thm:intro-bernstein-rect} we specify a cutoff $t = \sqrt{R}$ so that
\begin{align*}
\Prob{\max_{i \in [d]} \|\mtx{\Sigma}_2^{\frac{1}{2}} \mtx{U}_2^\mathsf{T} \mtx{\Pi}_{(i)}\| > t \defeq \frac13 u_2 R_2 + \sqrt{\frac19 u_2^2 R_2^2 + 2u_2 \sigma_2^2}} 
&\leq d (n - d_\delta + 1) \exp(-u_2) \\
&\leq \frac{\rho}{3}.
\end{align*}
We then have $u_2 \sim \log \frac{n}{\rho}$ and $R \sim u_2^2 R_2^2 + u_2 \sigma_2^2$.
We can verify the requirement equation~\eqref{eqn:sec_cond_R} is actually satisfied by the derived conditions on $m, d$ in equation~\eqref{eqn:k-satis-cond}.
\end{proof}

\subsection{Technical Lemmas}

We give the proof for the following technical lemma, which further transforms the bound in Theorem~\ref{thm:intro-chernoff}.

\begin{lemma}
\label{lem:chernoff_bd}
For all $\eta \geq 0$, we have $\frac{\econst^{\eta}}{(1+\eta)^{1 + \eta}} \leq \exp\paren{-\frac{\eta^2}{2 + \eta}}$.
Furthermore, for $\eta > 1$, we have $\exp\paren{-\frac{\eta^2}{2 + \eta}} \leq \exp\paren{-\frac{\eta^2}{3\eta}} \leq \exp\paren{-\frac{1}{3}\eta}$.
\end{lemma}

\begin{proof}
It suffices to show $(1 + \eta) \ln\paren{1 + \eta} - \eta \geq \frac{\eta^2}{2 + \eta}$ by taking the log of both sides. 
The new inequality can be obtained through the fact that
$$
({2 + \eta}) \ln\paren{1 + \eta} \geq 2 \eta, \quad \forall \eta \geq 0.
$$
We note for both sides of the inequality, 
$\ln(1+0) = 0 = \frac{2 \cdot 0}{2 + 0}$ and the derivative of $\ln(1+\eta)$ is always larger than the one of $\frac{2 \eta}{2 + \eta}$.
\end{proof}

\end{document}